\documentclass[sigconf,screen]{acmart}
\AtBeginDocument{%
  }

\setcopyright{acmlicensed}
\copyrightyear{2026}
\acmYear{2026}
\setcopyright{cc}
\setcctype{by-nc-nd}
\acmConference[MM '26]{Proceedings of the 35th ACM International Conference on Multimedia}{November 10--14, 2026}{Rio de Janeiro, Brazil}
\acmBooktitle{Proceedings of the 35th ACM International Conference on Multimedia (MM '26), November 10--14, 2026, Rio de Janeiro, Brazil}
\acmDOI{10.1145/3767308.3835106}
\acmISBN{979-8-4007-2213-4/2026/11}
\usepackage{xcolor}
\usepackage{tabularray}
\usepackage{pgf}
\usepackage{colortbl}
\usepackage{xcolor}
\usepackage{multirow}
\usepackage{geometry}
\usepackage{tabularx}

\definecolor{theme}{HTML}{50B4A8}

\newcommand{\cI}{\cellcolor{theme!45}}  
\newcommand{\cII}{\cellcolor{theme!30}} 
\newcommand{\cIII}{\cellcolor{theme!20}}
\newcommand{\cIV}{\cellcolor{theme!10}} 
\newcommand{\cV}{\cellcolor{theme!5}}   

\begin{document}

\title{When Efficiency Becomes Fragility: Exploiting Dynamic Routing Vulnerabilities in Adaptive UAV Tracking}

\author{Shaofeng Liang}
\email{shawnliang0420@gmail.com}
\affiliation{%
  \institution{The Hong Kong University of Science and Technology (Guangzhou)}
  \city{Guangzhou}
  \country{China}
}
\author{Runwei Guan}
\affiliation{%
  \institution{The Hong Kong University of Science and Technology (Guangzhou)}
  \city{Guangzhou}
  \country{China}
}
\author{Wenshuo Chen}
\affiliation{%
  \institution{The Hong Kong University of Science and Technology (Guangzhou)}
  \city{Guangzhou}
  \country{China}
}

\author{Jiemin Wu}
\affiliation{%
  \institution{The Hong Kong University of Science and Technology (Guangzhou)}
  \city{Guangzhou}
  \country{China}
}

\author{Bowen Tian}
\affiliation{%
  \institution{The Hong Kong University of Science and Technology (Guangzhou)}
  \city{Guangzhou}
  \country{China}
}

\author{Haozhe Jia}
\affiliation{%
  \institution{The Hong Kong University of Science and Technology (Guangzhou)}
  \city{Guangzhou}
  \country{China}
}

\author{Kaishen Yuan}
\affiliation{%
  \institution{The Hong Kong University of Science and Technology (Guangzhou)}
  \city{Guangzhou}
  \country{China}
}

\author{Songning Lai}
\affiliation{%
  \institution{The Hong Kong University of Science and Technology (Guangzhou)}
  \city{Guangzhou}
  \country{China}
}

\author{Daizong Liu}
\affiliation{%
  \institution{Wuhan University}
  \city{Wuhan}
  \country{China}
}

\author{Yutao Yue}
\authornote{Corresponding author}
\affiliation{%
  \institution{The Hong Kong University of Science and Technology (Guangzhou)}
  \city{Guangzhou}
  \country{China}\\
  \institution{Institute of Deep Perception Technology (JITRI)}
  \city{Wuxi}
  \country{China}
}
\renewcommand{\shortauthors}{Shaofeng Liang et al.}

\begin{abstract}
Resource constraints on UAV platforms have driven a paradigm shift in aerial tracking, from pursuing performance toward balancing accuracy with efficiency. Adaptive Transformer Trackers, which leverage an input-dependent dynamic routing architecture, have emerged as a representative solution to this challenge. However, we reveal that behind this computation-on-demand flexibility hides a critical structural flaw: the Lipschitz singularity of computational path decisions, which has an unbounded local Lipschitz constant at discrete layer-skipping decision boundaries. This mathematical discontinuity renders adaptive tracking networks inherently unstable: tiny input perturbations can be amplified at the gating modules, causing dramatic changes in the inference topology.  We formally characterize this singularity in the context of adaptive tracking architectures and, for the first time, identify it as a directly exploitable new attack surface. This insight reveals a previously overlooked and highly vulnerable topological path space attack surface. Unlike traditional adversarial attacks that target the output space, this new attack surface allows for the simultaneous manipulation of both the model’s semantic representation and its inference topology. Based on this, we propose the Adversarial Path-Inversion (API) framework. API generates imperceptible perturbations to precisely manipulate the gating decisions, forcing the inference onto altered computational paths. The severe inconsistency between the original and the inverted paths dismantles the representation capability of the model.  Extensive experiments on state-of-the-art adaptive trackers demonstrate that API achieves superior perturbation stealthiness, more effective attack, and faster inference speeds. This work opens a new dimension for the security analysis of dynamic tracking networks and provides a theoretical warning for constructing robust adaptive tracking architectures in the future.
\end{abstract}

\maketitle

\section{Introduction}
\begin{figure*}[!t]
    \centering
    \includegraphics[width=\linewidth]{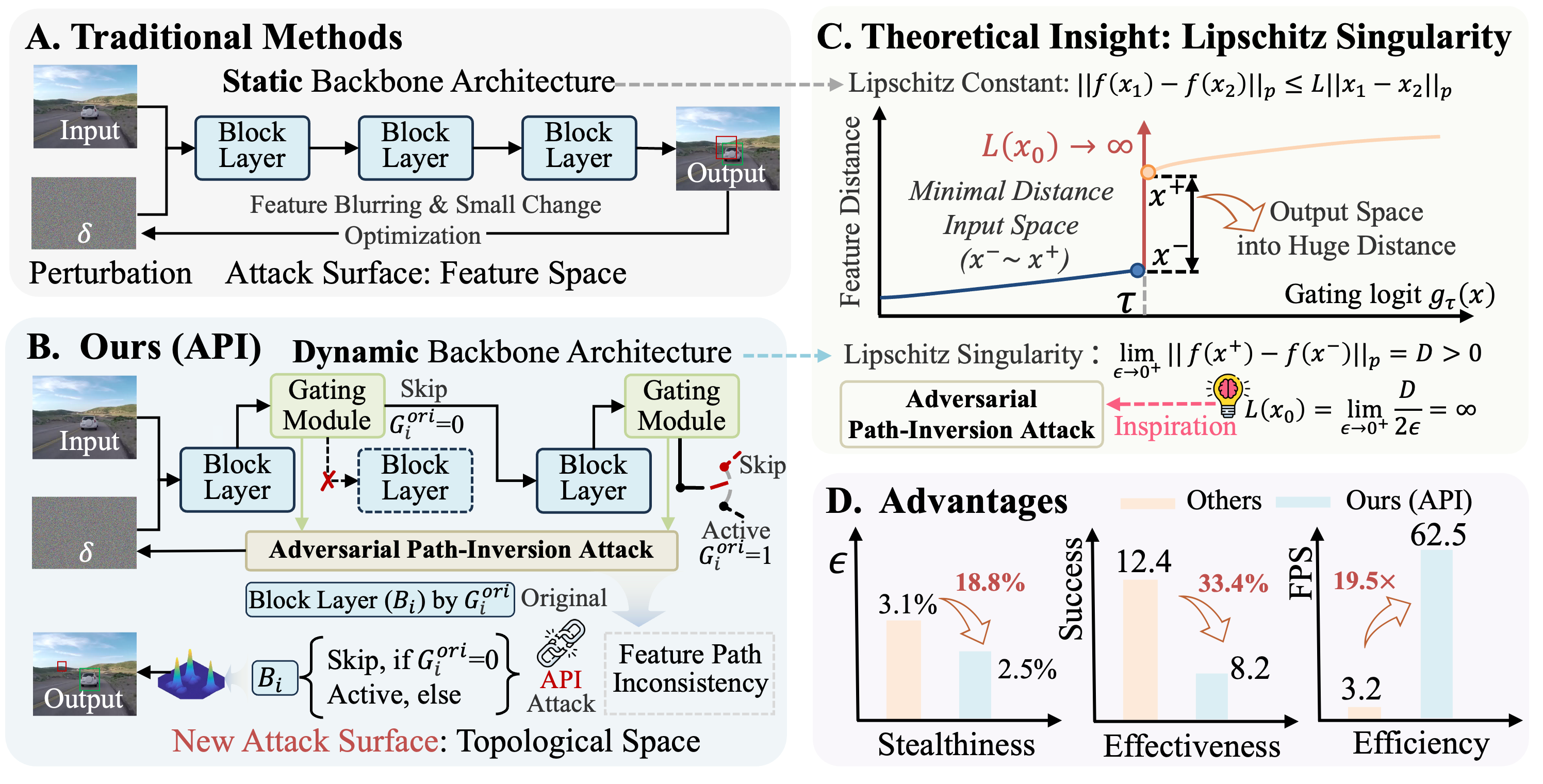}
    \vspace{-2.2em}
    \caption{Core motivation and theoretical foundation of the Adversarial Path-Inversion (API) framework. Unlike traditional attacks targeting the output space (subfig. A), API discovers attack surfaces in new topological spaces (subfig. B) and forces path changes by flipping gating decisions, leading to feature collapse. API theoretically proves the inherent instability of dynamic routing architectures (subfig. C). Compared to existing paradigms, API demonstrates superior perturbation stealthiness, more effective attack, and faster attack inference speeds (subfig. D).}
    \label{fig:motivation}
\end{figure*}
Unmanned Aerial Vehicles (UAVs) have been widely deployed in visual surveillance~\cite{aerialmind}, autonomous navigation~\cite{sun2026autofly}, and aerial media production~\cite{uavmedia}, where robust and real-time visual tracking plays an indispensable role.
In recent years, Vision Transformers (ViTs) have driven the performance of visual tracking~\cite{2021Transformer, OSTrack, TATrack,avt,liang2025cognitive} to new heights. However, the heavy computational~\cite{MixFormer} burden inherent in standard ViT architectures strictly limits the deployment of these models on resource-constrained platforms of UAVs. Consequently, the paradigm of UAV visual tracking is undergoing a critical shift from being purely performance-first to seeking a delicate balance between performance and efficiency. To address this issue, pioneering works have introduced Adaptive Vision Transformers~\cite{rao2021dynamicvit,yin2022vit}, represented by a prominent class of frameworks that includes AVTrack~\cite{avtrack}, SGLATrack~\cite{sglatrack}, LGTrack~\cite{LGTrack}, and DyTrack~\cite{dytrack}. By employing input-dependent layer-skipping mechanisms, such as dynamic activation modules, these computation-on-demand architectures can selectively activate transformer blocks. This dynamic routing allows the models to achieve real-time inference speeds while maintaining state-of-the-art tracking accuracy.

While these adaptive tracking models have achieved a remarkable balance between performance and efficiency, their adversarial robustness remains a conspicuously under-explored frontier. For unmanned aerial vehicles (UAVs) operating in dynamic environments, this can seriously hinder their reliable physical deployment and even physical collisions in high-speed flight scenarios. Existing adversarial attacks~\cite{jia2021iou,fu2022ad,xiang2025acattack,lai2024cat} in visual tracking predominantly target static neural network architectures. These attacks typically focus on perturbing to corrupt semantic representations or bounding box predictions within a fixed computational graph, as shown in Fig.~\ref{fig:motivation}-A. However, adaptive models introduce a fundamentally different computational paradigm, where the computational path itself depends on the input (Fig.~\ref{fig:motivation}-B).  This raises a natural question:  \textbf{Does such dynamic routing introduce new vulnerabilities beyond those from static tracking networks?}

To investigate this, we analyze that layer-skipping decisions in adaptive trackers are inherently discrete. Specifically, each gating module produces a binary output that either activates or bypasses a Transformer block.  This discrete switching behavior implies that the output of the network may exhibit discontinuities in the vicinity of the gating boundaries. To formalize this intuition, we evaluate the Lipschitz continuity (\emph{i.e}, the sensitivity of the outputs to inputs) of the gating decisions, and derive the local Lipschitz singularity at these discrete decision boundaries (see Fig.~\ref{fig:motivation}-C).  This analysis reveals that the resulting singularity indirectly constitutes a directly exploitable attack surface in adaptive tracking, leading to a concrete security consequence: the discrete nature of layer-skipping decisions means that even infinitesimal input perturbations can trigger abrupt transitions in the inference topology, forcing the computational graph to deviate from its intended execution path. This finding exposes a previously overlooked threat category that we term the Topology-Path-Based Attack. Traditional adversarial attacks~\cite{fu2022ad,xiang2026adversarial,wang2025fa3t} operate within a fixed computational path and can only degrade the semantic representation along that path. In contrast, the attack surface identified here enables an adversary to simultaneously alter both the dynamic inference topology and the feature representation, resulting in a far more severe and irreversible collapse of tracking performance.

To systematically exploit this vulnerability, we propose the Adversarial Path-Inversion (API) framework. By optimizing perturbations specifically tailored for the gating modules of adaptive trackers, the proposed framework precisely manipulates the discrete routing decisions to force the inversion of computational paths. It dismantles the representation capability of the tracker by exploiting the severe feature inconsistency between the original optimal path and the maliciously altered computational paths. Remarkably, we discover that directly targeting the inference topology inflicts a significantly more severe degradation on tracking precision compared to traditional attacks. This finding strongly validates the effectiveness and novelty of the discovered topology-path-based attack surface. Comprehensive evaluations across state-of-the-art adaptive trackers substantiate that the API framework achieves superior perturbation stealthiness, heightened attack efficacy, and accelerated inference speeds (Fig.~\ref{fig:motivation}-D). Furthermore, it exhibits remarkable generalizability, consistently achieving high attack efficacy across a broad class of adaptive architectures.  This work unveils a novel dimension for the security assessment of dynamic tracking models and establishes a critical theoretical imperative for the development of robust adaptive architectures in the future.

In summary, this work makes the following contributions:

(1) We analyze the inherent instability of dynamic routing in adaptive tracking architectures from a Lipschitz perspective, showing that discontinuous gating decisions can lead to vulnerabilities around routing boundaries.

(2) We expose a previously overlooked attack surface rooted in the computational path and propose the Adversarial Path-Inversion framework, which disrupts dynamic trackers by maliciously inverting their routing decisions.

(3) Extensive experiments across multiple adaptive trackers and  UAV benchmarks demonstrate that the proposed framework significantly outperforms traditional output-oriented attacks in degrading tracking precision.
\vspace{-1em}
\section{Related Work}
\subsection{Visual Object Tracking}
Visual object tracking is a fundamental task in computer vision, aiming to continuously estimate the state and location of a target across subsequent frames given its initial state. While early methods based on discriminative correlation filters~\cite{kcf,srdcf,DIMP,atom} and Siamese networks~\cite{SiamBAN,SiamFC++,SiamCAR} established crucial baselines, their reliance on local operations limits the capability to capture global context. Recently, Vision Transformer (ViT)-based trackers~\cite{2021Transformer,TATrack}, such as MixFormer~\cite{MixFormer} and OSTrack~\cite{OSTrack}, have significantly enhanced feature discriminability by enabling global interactive modeling between the template and the search region. However, their heavy computational burden poses a severe deployment challenge on resource-constrained hardware platforms, such as Unmanned Aerial Vehicles.
To overcome this, the research community has shifted attention towards efficient visual tracking frameworks based on dynamic Transformers~\cite{rao2021dynamicvit,yin2022vit,kang2023exploring}. The core idea of these dynamic architectures is to allocate computational resources on demand based on the complexity of the input instance.  These models~\cite{avtrack,dytrack} insert multiple decision makers into the intermediate layers of the ViT to evaluate the reliability of the current feature representation. Similarly, SGLATrack \cite{sglatrack} employs a selection module acting as a decision maker, which outputs the selection probability of subsequent layers and retains only the optimal representative layer from deep stages. Despite their remarkable efficiency, we reveal that these decision-based routing mechanisms inadvertently expose a fatal structural vulnerability~\cite{slowformer,Slowdown}, creating massive hidden dangers regarding adversarial robustness, a previously unexplored blind spot that our work directly targets. 
\vspace{-1em}
\subsection{Adversarial Attacks in Visual Tracking}
With the widespread application of tracking algorithms in safety-critical domains such as autonomous driving and surveillance systems, evaluating their robustness against malicious attacks has become increasingly urgent. Adversarial attacks~\cite{chakraborty2021survey,du2025comprehensive,du2025transfficformer,lai2024guarding} trick deep neural networks into making incorrect predictions by injecting imperceptible perturbations into input images. Existing adversarial perturbation methods in visual tracking can be broadly categorized into online iterative optimization methods \cite{online1,trackpgd,rtaa,jia2021iou} and offline generator-based paradigms \cite{ooa,TASF,csa}. Online methods, such as RTAA \cite{rtaa}, generate strong adversarial examples by utilizing pseudo-labels to progressively align the correct classification and regression labels of the tracker. While online optimization brings significant improvements in attack performance, its time-consuming nature heavily compromises stealthiness, especially in high-speed tracking scenarios. In contrast, offline methods~\cite{xiang2026adversarial} train an adversarial generator from massive data, requiring only a single forward pass to generate perturbations during the attack phase, thereby achieving an optimal balance between time and performance. For instance, the Only Once Attack framework \cite{ooa} effectively fools the tracker by generating perturbations solely for the initial template.
Furthermore, IoU Attack \cite{jia2021iou} introduces a decision-based black-box approach that sequentially generates perturbations based on predicted IoU scores. Ad2Attack \cite{fu2022ad} reveals a neglected vulnerability from the perspective of image preprocessing, targeting the image resampling process. More recently, with the rise of multi-modal tracking, researchers have proposed sophisticated cross-modal attacks. For example, FA3T \cite{wang2025fa3t} disrupts cross-modal alignment through feature-aware perturbations; ACAttack \cite{xiang2025acattack} achieves adaptive cross attacks via multi-modal response decoupling; and methods like ICAttack \cite{xiang2026adversarial} and Cross-Modal Stealth \cite{xiang2025cross} progressively dismantle fusion mechanisms via intra-modal excavation and coarse-to-fine latent space strategies. 

However, all the aforementioned attacks predominantly focus on adding perturbations at the semantic feature or input pixel level to strictly corrupt the final output. Unlike these feature-level attacks, our method pioneers the manipulation of dynamic routing decisions, revealing that attacking the inference topology is far more devastating. Some related studies~\cite{slowformer, Slowdown} have also explored adversarial attacks targeting dynamic network routing mechanisms in image classification. These attacks focus on inference efficiency in image classification, rather than the visual tracking task, without considering the crucial temporal dynamics, template search relevance, and online state estimation inherent in visual tracking, thus serving a fundamentally different attack objective and task domain.

\section{Method}
\subsection{Problem Formulation}
Visual object tracking aims to continuously estimate the state of a target in a search region sequence $S = \{s_1, s_2, \dots, s_T\}$, given an initial template $z$. A typical tracker can be formulated as a mapping function $\mathcal{F}$, which predicts the bounding box $b_t$ at frame $t$ via $b_t = \mathcal{F}(z, s_t; \Theta)$, where $\Theta$ represents the model parameters. Referring to standard adversarial attack paradigms in visual tracking \cite{ooa, wang2025fa3t, rtaa}, the goal of an adversary is to craft an imperceptible perturbation $\delta$, added to the input images to induce tracking failure. Formally, the traditional objective is to maximize the deviation between the predicted bounding box of the adversarial sample and the ground truth, formulated as:
\begin{equation}
    \label{eq:standard_attack}
    \max_{\delta} \mathcal{L}_{task}(\mathcal{F}(z, s_t + \delta; \Theta), y_t), \quad \text{s.t.} \quad |\delta| \leq \epsilon
\end{equation}
where $y_t$ is the ground truth label, $\mathcal{L}_{task}$ is the tracking loss, and $\epsilon$ is the maximum perturbation magnitude constrained to ensure visual stealthiness. 

\begin{figure*}[!t]
	\centering
	\includegraphics[width=\linewidth]{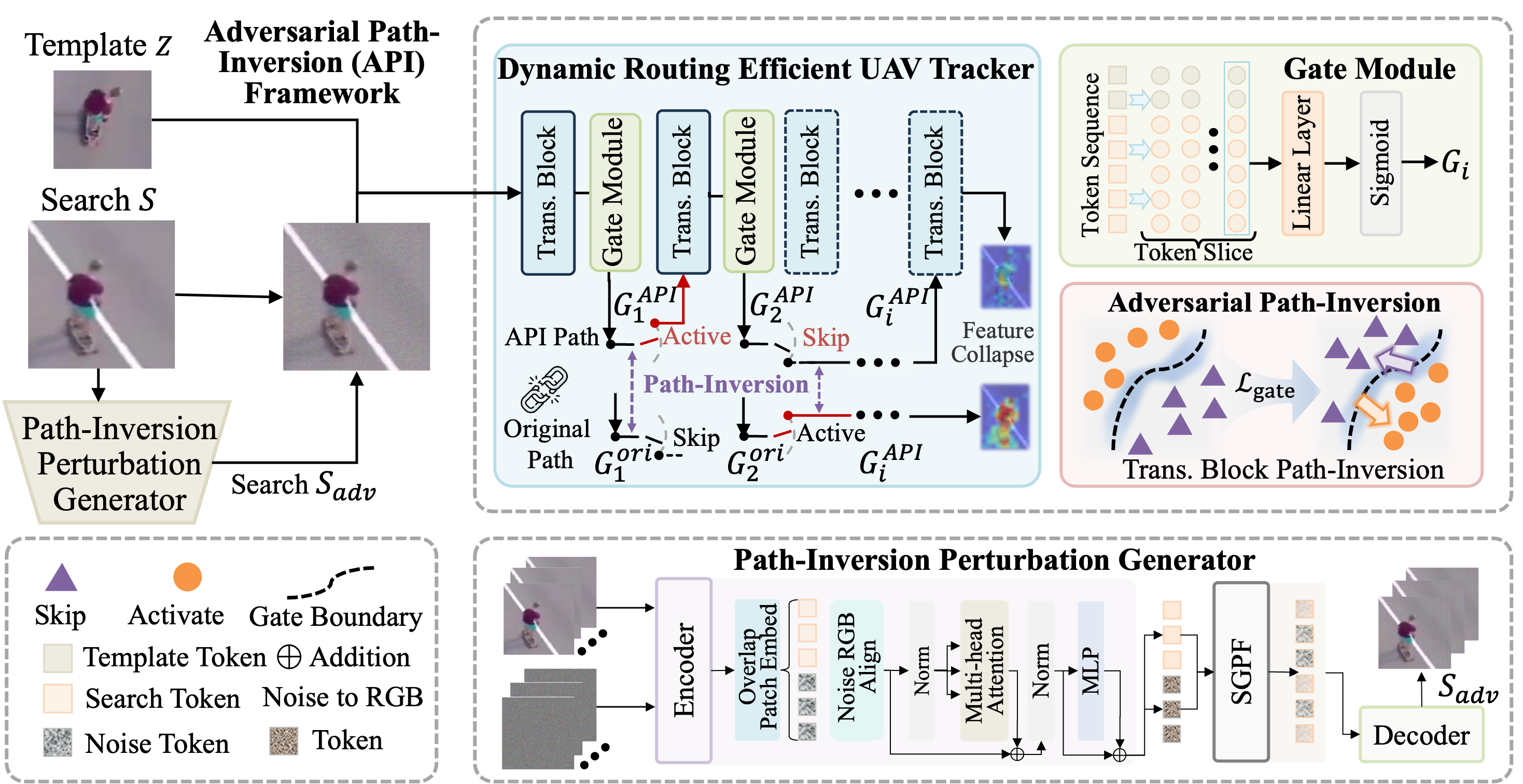}
	\hfill
    \vspace{-2em}
	\caption{Overview of the Adversarial Path-Inversion (API) framework. The perturbation generator (bottom) synthesizes adversarial noise from the search region and a learnable noise prior, which is injected before patch embedding to invert the gating decisions of the victim tracker (top), dismantling feature consistency across computational paths.}
	\label{fig:pipeline}
    \vspace{-1em}
\end{figure*}
\subsection{Theoretical Analysis: The Lipschitz Singularity in Dynamic Routing}

Adaptive tracking architectures achieve high efficiency by introducing intermediate decision modules, such as dynamic gating mechanisms, to selectively skip redundant transformer blocks. To formally characterize the resulting system, let $x \in \mathbb{R}^n$ denote the intermediate feature representation prior to a routing module. The decision function $\pi(\cdot)$ can be abstracted as a discrete step function:\begin{equation}\pi(x) = \mathbb{I}(g(x) > \tau)\end{equation}where $g(\cdot)$ represents a scoring network (e.g., the activation module in AVTrack~\cite{avtrack} or the selection module in SGLATrack~\cite{sglatrack} and LGTrack~\cite{LGTrack}), $\tau$ is a predefined confidence threshold, and $\mathbb{I}(\cdot)$ is the indicator function. 

Depending on the binary value of $\pi(x) \in \{0, 1\}$, the network activates divergent computational sub-graphs, denoted as $f_{act}$ and $f_{skip}$. The resulting output feature mapping can be formulated as a piecewise discontinuous function:
\begin{equation}
f(x) = \pi(x) f_{act}(x) + (1 - \pi(x)) f_{skip}(x)
\end{equation}

While this mechanism provides remarkable flexibility for computation-on-demand, we reveal that it induces severe structural instability. This vulnerability can be formalised by analyzing the Lipschitz continuity of the underlying mapping. A mapping is defined as $L$-Lipschitz continuous if, for any inputs $x_1, x_2$, the following condition holds:
\begin{equation}\|f(x_1) - f(x_2)\|_p \le L \|x_1 - x_2\|_p\end{equation}
where $L$ is the Lipschitz constant, $\|\cdot\|_p$ represents the $L_p$ distance. 

The discrete nature of the gating mechanism induces a structural singularity at the decision boundary. To formalize this, consider an input $x_0$ residing on the decision boundary manifold $\mathcal{D}_b$. We define a vanishingly small directional perturbation $\delta$ and two neighboring points $x^+$ and $x^-$ on opposite sides of the manifold:
\begin{equation}
\begin{aligned}
\mathcal{D}_b &= \{x \in \mathbb{R}^n : g(x) = \tau\}\\
x^{\pm} &=x_0 \pm \delta = x_0 \pm \epsilon \cdot \vec{u} \quad \text{s.t.} 
\quad g(x^+) > \tau, \ g(x^-) \le \tau
\end{aligned}
\end{equation}
where $\vec{u}$ is a unit vector transversal to the boundary and $\epsilon \to 0^+$ is an infinitesimal scalar. Although the distance between these points in the input space vanishes as $\epsilon$ approaches zero,  the discrepancy between the corresponding outputs remains finite and non-zero:
\begin{equation}
\lim_{\epsilon \to 0^+} \|f(x^+) - f(x^-)\|_p = \|f_{act}(x_0) - f_{skip}(x_0)\|_p = D > 0
\end{equation}

The local Lipschitz constant $L(x_0)$ is defined as the supremum of the ratio between the output change and the input perturbation. By evaluating this ratio across the boundary, we obtain:
\begin{equation}
L(x_0) = \lim_{\epsilon \to 0^+} \frac{\|f(x^+) - f(x^-)\|_p}{\|x^+ - x^-\|_p} \ge \lim_{\epsilon \to 0^+} \frac{D}{2\epsilon} = \infty
\end{equation}

where $L(x_0)$ becomes unbounded and approaches infinity. 

The above analysis reveals that the discrete gating mechanism introduces local discontinuities, making routing decisions highly sensitive to tiny perturbations and easy to invert. In adaptive tracking architectures, different routing paths correspond to distinct sub-networks, leading to semantically inconsistent feature representations. This inconsistency is particularly harmful for visual tracking, where accurate localization relies on the precise alignment between template and search features across consecutive frames. Once disrupted, the resulting misalignment quickly accumulates, causing irreversible tracking drift.

\subsection{Adversarial Path-Inversion Framework}
Motivated by the theoretical discovery of the above, we propose the Adversarial Path-Inversion (API) framework to systematically exploit this vulnerability and manipulate the dynamic routing decisions of adaptive trackers, as illustrated in Fig.~\ref{fig:pipeline}. 

The inputs $z \in \mathbb{R}^{H_z \times W_z \times 3}$ denote the initial template image, which provides the reference appearance, and let $s_t \in \mathbb{R}^{H_x \times W_x \times 3}$ represent the search region at frame $t$. The tracking task can be formulated as learning a mapping function $\mathcal{F}(\cdot)$ to predict the bounding box $B_t$ of the target, where the global mapping is essentially a composition of $L$ Transformer blocks $\mathcal{B}_l$, each conditionally executed based on a discrete gate $g_{l} \in \{0, 1\}$:
\begin{equation}
\begin{aligned}
B_t = \mathcal{F}(z, s_t) &= \mathcal{H} \Big( \Phi_L(\Phi_{L-1}(\cdots\Phi_1(\mathcal T_0))) \Big), \\
\Phi_l(x) &= g_{l} \mathcal{B}_l(x) + (1 - g_{l})x    
\end{aligned}
\end{equation}
where $\mathcal{H}(\cdot)$ denotes the tracking head,  and $\mathcal{T}_0$ represents the initially embedded joint tokens.

The API framework employs an encoder-decoder perturbation generator $G_p$ to synthesize a localized adversarial noise $\delta_t$. This perturbation is strictly applied to the search region prior to the patch embedding (PE) stage:
\begin{equation}
\tilde{x}_t = x_t + \delta_t = x_t + G_p(x_t), \quad \text{s.t.} \quad \delta_t \leq \epsilon
\end{equation}
where $\epsilon$ denotes the maximum magnitude of the perturbation. The clean template $z$ and the perturbed search input $\tilde{x}_t$ are subsequently processed by the patch embedding module of the victim tracker to obtain the initial joint tokens $\mathcal{T}_{0}^{adv}$:
\begin{equation}
\mathcal{T}_{0}^{adv} = [\text{PE}(z); \text{PE}(\tilde{x}_t)] + \mathcal{P}_{pos}
\end{equation}
where $[\cdot]$ denotes the concatenation operation, and $\mathcal{P}_{pos}$ represents the position embeddings.  The ultimate goal is to force the current routing decisions $\{r^1_{t}, \dots, r^L_{t}\}$ to deviate from the clean computational path, thereby dismantling the feature consistency and the overall tracking precision. During training, the parameters of the victim tracker remain frozen, and only the generator is optimized.

\textbf{Path-Inversion Perturbation Generator.}
To systematically exploit the structural vulnerabilities inherent in adaptive trackers, we design a gate-aware adversarial perturbation generator. Drawing inspiration from the intra-modal excavation paradigm~\cite{xiang2026adversarial}, the framework processes the search region $s_t$ and the noise prior $n$ through two independently parameterized branches. These branches utilize a hierarchical encoding architecture to progressively extract features, yielding the scene tokens $\mathcal{T}_s$ and noise tokens $\mathcal{T}_n$:
\begin{equation}
\mathcal{T}_s = \mathcal{E}^{(s)}(s_t),
\quad
\mathcal{T}_n = \mathcal{E}^{(n)}(n),
\quad
\mathcal{T}_s, \mathcal{T}_n \in \mathbb{R}^{N \times C},
\label{eq:encoder_unified}
\end{equation}
where $\mathcal{E}^{(s)}$ and $\mathcal{E}^{(n)}$ denote the scene and noise encoders.

Subsequently, to achieve precise spatial targeting and ensure visual stealthiness, we incorporate an Saliency-Guided Perturbation Focusing (SGPF) module and an adversarial perturbation decoder $\mathcal{D}(\cdot)$. The SGPF module identifies decision-critical tokens to concentrate adversarial energy, while the decoder translates these latent representations into a localized pixel-level perturbation $\delta_t$. This process is summarized by the following joint mapping:
\begin{equation}
\mathcal{T}^{\mathrm{tc}} = \mathrm{SGPF}(\mathcal{T}_s + \mathcal{T}_n),
\quad
\delta_t = \mathcal{D}(\mathcal{T}^{\mathrm{tc}}, \mathcal{T}_s),
\label{eq:joint_mapping_unified}
\end{equation}
where $\mathcal{T}^{\mathrm{tc}}$ represents the token-centric adversarial features. Through this mechanism, abstract topological threats are materialized into scene-adaptive noise, the specific implementations of which are detailed in the following sections.

As the cornerstone of our framework, we formulate the Path-Inversion Objective ($\mathcal{L}_{\mathrm{path}}$). Let $l_{t,k}^{\mathrm{adv}}$ and $l_{t,k}^{\mathrm{clean}}$ denote the pre-sigmoid logits of the $k$-th gating module at frame $t$ for the adversarial and clean inputs, respectively. We define a polarity indicator $\eta_{t,k} = \mathrm{sgn}(l_{t,k}^{\mathrm{clean}})$, where $\eta_{t,k} = 1$ for active layers and $\eta_{t,k} = -1$ for skipped layers. The objective is:
\begin{equation}
\mathcal{L}_{\mathrm{path}} = \frac{1}{T \cdot K} \sum_{t=1}^{T} \sum_{k=1}^{K} \eta_{t,k} \cdot l_{t,k}^{\mathrm{adv}}.
\label{eq:gate_loss}
\end{equation}
where $T=9$ is the number of frames in each training segment. Minimizing $\mathcal{L}_{\mathrm{path}}$ drives $l_{t,k}^{\mathrm{adv}}$ toward the opposite sign of its clean counterpart: originally active layers are forced to skip, and originally skipped layers are forced to activate, thereby systematically inverting the computational path.

To complement the topological disruption, we introduce a Synergistic Feature Disruption objective ($\mathcal{L}_{\mathrm{feat}}$) that minimizes the cosine similarity between clean and adversarial token embeddings:
\begin{equation}
\mathcal{L}_{\mathrm{feat}} = \frac{1}{T} \sum_{t=1}^{T} \frac{1}{N_s} \sum_{i=1}^{N_s} \mathrm{sim}_{\mathrm{cos}}(f_{t,i}^{\mathrm{adv}}, f_{t,i}^{\mathrm{clean}}),
\label{eq:feat_loss}
\end{equation}
where $N_s$ denotes the number of search tokens. This objective drives the semantic representation away from its clean counterpart, amplifying the destructive impact of path inversion.

\textbf{Saliency-Guided Perturbation Focusing.}
To concentrate adversarial energy on decision-critical spatial regions, inspired by~\cite{wang2025fa3t, OSTrack,xiang2026adversarial}, we introduce the Saliency-Guided Perturbation Focusing (SGPF) module. After the final encoder stage, the scene and noise tokens are fused into a unified representation $\mathcal{T}' = \mathcal{T}_s + \mathcal{T}_n$. We compute the self-attention map over $\mathcal{T}'$ and average to obtain a per-token relevance score. The top-$K$ tokens with the highest scores form a binary saliency mask $W \in \{0,1\}^N$. The adversarial representation is then modulated as:
\begin{equation}
\mathcal{T}^{\mathrm{tc}} = W \odot (\mathcal{T}') + \gamma(1 - W) \odot \mathcal{T}_n, \quad \gamma = 0.5.\label{eq:SGPF_modulation}
\end{equation}
For salient tokens ($W_i = 1$), the full adversarial signal augmented with the noise prior is preserved, directing maximum perturbation energy toward the regions that dominate the gating decisions of the tracker. For non-salient tokens ($W_i = 0$), only a scaled noise component is retained ($\gamma = 0.5$), which suppresses perturbation magnitude in background regions and improves visual imperceptibility.
\begin{table*}[t]
\centering
\renewcommand{\arraystretch}{1.2} 
\setlength{\tabcolsep}{4pt} 
\caption{Quantitative comparison of adversarial attack methods on AVTrack across UAV tracking benchmarks. The top 5 best attacking results in each metric are highlighted with decreasing color depths. UAV123* denotes the UAV123@10fps subset.}
\label{tab:attack_performance}
\begin{tabularx}{\linewidth}{ l *{12}{>{\centering\arraybackslash}X} }
\toprule
\multirow{2}{*}{\textbf{Method}} & 
\multicolumn{2}{c}{\textbf{DTB70}} & 
\multicolumn{2}{c}{\textbf{UAV123}} & 
\multicolumn{2}{c}{\textbf{UAVDT}} & 
\multicolumn{2}{c}{\textbf{VisDrone}} & 
\multicolumn{2}{c}{\textbf{UAV123*}} & 
\multicolumn{2}{c}{\textbf{UAVtrack 112}} \\
\cmidrule(lr){2-3} \cmidrule(lr){4-5} \cmidrule(lr){6-7} \cmidrule(lr){8-9} \cmidrule(lr){10-11} \cmidrule(lr){12-13}
 & 
Pres. & Succ. & 
Pres. & Succ. & 
Pres. & Succ. & 
Pres. & Succ. & 
Pres. & Succ. & 
Pres. & Succ. \\
\midrule
AVTrack~\cite{avtrack}   & 84.3 & 65.0 & 84.8 & 66.8 & 82.1 & 58.7 & 86.0 & 65.0 & 83.2 & 65.8 & 80.3 & 65.4 \\
\midrule
Random        & 82.3 & 61.8 & 82.7 & 64.6 & 76.3 & 52.9 & 83.2 & 62.6 & 81.7 & 64.0 & 79.2 & 63.9 \\
FGSM~\cite{fgsm}    & \cV 53.3 & \cIV 34.7 & \cV 60.8 & \cV 43.5 & 56.3 & 41.5 & \cV 50.0 & \cIV 26.3 & \cIV 56.0 & \cIV 38.3 & 57.0 & 39.5 \\
UAP~\cite{uap}      & \cIV 48.2 & \cV 35.8 & \cIV 54.3 & \cIV 41.5 & \cIV 31.0 & \cIV 20.3 & \cIV 43.0 & \cV 32.1 & \cV 57.4 & \cV 43.7 & \cIV 50.0 & \cIV 37.6 \\
CSA~\cite{csa}             & 71.2 & 47.6 & 74.0 & 49.8 & 62.8 & 35.5 & 72.5 & 52.5 & 69.9 & 48.2 & 68.5 & 46.2 \\
TTAttack~\cite{ttattack}             & 54.0 & 35.9 & 67.2 & 46.8 & \cV 45.6 & \cV 23.6 & 65.8 & 43.0 & 65.9 & 46.8 & \cV 54.8 & \cV 37.9\\
DFA~\cite{dfa}       & 72.3 & 55.1 & 72.8 & 57.2 & 71.9 & 49.0 & 78.6 & 58.8 & 73.1 & 57.6 & 71.4 & 56.5 \\
Ad2attack~\cite{fu2022ad}  & 71.9 & 48.4 & 78.5 & 57.6 & 67.2 & 41.8 & 76.7 & 54.7 & 75.6 & 56.1 & 73.3 & 54.4 \\
RTAA~\cite{rtaa}      & \cII 19.0 & \cII 12.4 & \cII 27.3 & \cII 20.4 & \cIII 25.3 & \cIII 16.8 & \cIII 33.0 & \cIII 24.2 & \cII 34.2 & \cII 25.0 & \cII 21.2 & \cII 16.0 \\
ICAttack~\cite{xiang2026adversarial}   & \cIII 38.0 & \cIII 30.3 & \cIII 38.9 & \cIII 31.7 & \cII 20.8 & \cII 15.1 & \cII 24.8 & \cII 19.0 & \cIII 41.3 & \cIII 33.5 & \cIII 34.2 & \cIII 27.2 \\
\textbf{API Attack} & \cI \textbf{16.1} & \cI \textbf{8.2} & \cI \textbf{11.6} & \cI \textbf{7.5} & \cI \textbf{18.2} & \cI \textbf{8.5} & \cI \textbf{16.4} & \cI \textbf{7.0} & \cI \textbf{16.1} & \cI \textbf{10.2} & \cI \textbf{16.8} & \cI \textbf{10.4} \\
\bottomrule
\end{tabularx}
\end{table*}

\textbf{Adversarial Perturbation Decoder.}
The decoder translates the adversarial token representation $\mathcal{T}^{\mathrm{tc}}$ into a pixel-level perturbation while preserving multi-scale spatial coherence through encoder skip connections. It progressively upsamples through $S=3$ stages, where each stage $j$ doubles the spatial resolution via transposed convolution, fuses the result with the corresponding encoder feature, and refines it through a Transformer block:
\begin{equation}
F^j_{\mathrm{dec}} = \mathcal{B}^{(d)}_j \left( \mathrm{Up}_j ( F^{j-1}_{\mathrm{dec}} ) + F^{E}_{S-j} \right), \quad j = 1, 2, 3
\end{equation}
where $F^0_{\mathrm{dec}} = \mathcal{T}^{\mathrm{tc}}$, $\mathrm{Up}_j$ denotes transposed convolution with bilinear interpolation, $F^{E}_{S-j}$ is the skip feature from the encoder, and $\mathcal{B}^{(d)}_j$ is a Transformer block. A final upsampling followed by a $1 \times 1$ convolution and a $\tanh$ activation yields the perturbation:
\begin{equation}
\delta_t = \epsilon \cdot \tanh \left( \mathrm{Conv}_{1 \times 1} ( \mathrm{Up}_4 ( F^3_{\mathrm{dec}} ) ) \right)
\end{equation}
This formulation inherently enforces $\|\delta_t\|_\infty \leq \epsilon$ without external clipping. The final adversarial search frame is constructed as $s^{\mathrm{adv}}_t = \mathrm{clamp}(s_t + \delta_t, 0, 1)$.

Finally, to dismantle tracking precision at the output level, we incorporate a Response Center Suppression objective ($\mathcal{L}_{\mathrm{resp}}$). Inspired by \cite{xiang2025acattack,xiang2026adversarial}, we aggregate multi-frame segments during training to capture the temporal dynamics of tracking. Specifically, within a temporal window from $t$ to $t+T$, we suppress the response map toward a uniform distribution and simultaneously apply a Visual Reconstruction Constraint ($\mathcal{L}_{\mathrm{recon}}$) to preserve visual stealthiness:
\begin{equation}
    \mathcal{L}_{resp} = \frac{1}{T}\sum_{i=t}^{t+T-1} \|R^{adv}_i - \hat{R}_i\|_1, 
    \quad 
    \mathcal{L}_{\mathrm{recon}} = \frac{1}{T} \sum_{i=t}^{t+T-1} \| \hat{x}_i - x_i \|_1,
    \label{eq:temporal_losses}
\end{equation}
where $R_i^{\mathrm{adv}}$ is the adversarial response map at frame $i$ and $T$ is the temporal segment length. This multi-frame formulation ensures persistent attack efficacy across consecutive frames while satisfying strict stealthiness constraints.

\textbf{Path-Inversion Centric Composite Optimization.} The generator is optimized with a composite objective that jointly drives topological inversion, semantic disruption, and visual fidelity:
\begin{equation}
\mathcal{L}_{\mathrm{total}} = \lambda_{\mathrm{path}}\mathcal{L}_{\mathrm{path}} + \lambda_{\mathrm{resp}}\mathcal{L}_{\mathrm{resp}} + \lambda_{\mathrm{feat}}\mathcal{L}_{\mathrm{feat}} + \lambda_{\mathrm{recon}}\mathcal{L}_{\mathrm{recon}},
\label{eq:total_loss}
\end{equation}
where $\lambda$ denotes the designated scaling coefficients to balance the optimization landscape.

\section{Experiments}

\subsection{Experimental Settings}
\textbf{Implementation Details.} 
In our experiments, we employ state-of-the-art adaptive trackers AVTrack~\cite{avtrack}, SGLATrack~\cite{sglatrack}, and LGTrack~\cite{LGTrack} as the primary baseline victim model. The entire training process is conducted on the training split of the LaSOT~\cite{fan2021lasot} dataset. We utilize a computing cluster equipped with 4 NVIDIA RTX 4090 GPUs to train 50 epochs. During the training phase, the parameters of the baseline tracking model are strictly frozen, and the optimization process is exclusively directed toward the proposed Path-Inversion Perturbation Generator and the learnable noise seed. 
To ensure the visual imperceptibility of the attack, the maximum magnitude of the adversarial perturbation is strictly bounded to 0.025 in the normalized continuous pixel space.  Furthermore, the composite loss hyperparameters are empirically configured as follows: the path-inversion loss weight $\lambda_{path} = 2.0$, the response suppression weight $\lambda_{resp} = 1\times 10^4$, the feature disruption weight $\lambda_{feat} = 1.0$, and the reconstruction constraint weight $\lambda_{recon} = 1.0$.
\begin{table}[t]
\centering
\small 
\caption{Comparison of performance and efficiency degradation rates on the DTB70 dataset.}
\vspace{-1em}
\label{tab:efficiency_comparison}
\renewcommand{\arraystretch}{1.1} 
\begin{tabularx}{\linewidth}{ l c *{4}{>{\centering\arraybackslash}X} }
\toprule
\multirow{2}{*}{\textbf{Method}} & \multirow{2}{*}{\textbf{$\epsilon$(\%)}} & \multicolumn{2}{c}{\textbf{Performance Drop}} & \multicolumn{2}{c}{\textbf{Efficiency}} \\
\cmidrule(lr){3-4} \cmidrule(lr){5-6}
& & Pres.  & Succ. & FPS & Drop (\%) \\
\midrule
Victim~\cite{avtrack} & - & - & - & 79.5 & - \\
\midrule
DFA~\cite{dfa}               & 3.1 & 14.3 & 15.2 & 13.7 & 82.8 \\
Ad2attack~\cite{fu2022ad}       & 3.1 & 14.7 & 25.6 & \cII 49.7 & \cII 37.5 \\
CSA~\cite{csa}               & 3.1 & 15.6 & 26.9 & \cV 17.0 & \cV 78.6 \\
TTAttack~\cite{ttattack}       & 3.1 & 35.9 & 44.8 & \cIII 29.7 & \cIII 62.6 \\
Fgsm~\cite{fgsm}            & 3.1 & \cV 36.7 & \cIV 46.6 & 11.2 & 85.9 \\
UAP~\cite{uap}               & 3.1 & \cIV 42.9 & \cV 44.9 & 11.4 & 85.7 \\
ICattack~\cite{xiang2026adversarial}         & 2.5 & \cIII 54.9 & \cIII 53.3 & \cIV 22.8    & \cIV 71.3     \\
RTAA~\cite{rtaa}            & 3.1 & \cII 77.5 & \cII 81.0 & 4.6  & 94.2 \\
\textbf{API Attack} & \textbf{2.5} & \cI \textbf{80.9} & \cI \textbf{87.3} & \cI \textbf{62.5} & \cI \textbf{21.4} \\
\bottomrule
\end{tabularx}
\end{table}

\begin{table}[t]
\centering
\caption{Generalization Analysis of the API Framework.}
\vspace{-1em}
\label{tab:different_model}
\renewcommand{\arraystretch}{1.1}
\setlength{\tabcolsep}{1.5pt} 
\resizebox{\linewidth}{!}{%
\begin{tabular}{l c c c c c c c c c c c c}
\toprule
\multirow{2}{*}{\textbf{Tracker}} & \multicolumn{2}{c}{\textbf{DTB70}} & \multicolumn{2}{c}{\textbf{UAV123}} & \multicolumn{2}{c}{\textbf{UAVDT}} & \multicolumn{2}{c}{\textbf{VisDrone}} & \multicolumn{2}{c}{\begin{tabular}{@{}c@{}}\textbf{UAV123}\\\textbf{@10fps}\end{tabular}} & \multicolumn{2}{c}{\begin{tabular}{@{}c@{}}\textbf{Uavtrack}\\\textbf{112}\end{tabular}} \\
\cmidrule(lr){2-3} \cmidrule(lr){4-5} \cmidrule(lr){6-7} \cmidrule(lr){8-9} \cmidrule(lr){10-11} \cmidrule(lr){12-13}
 & Prec. & Succ. & Prec. & Succ. & Prec. & Succ. & Prec. & Succ. & Prec. & Succ. & Prec. & Succ. \\
\midrule
AVTrack~\cite{avtrack} & 84.3 & 65.0 & 84.8 & 66.8 & 82.1 & 58.7 & 86.0 & 65.0 & 83.2 & 65.8 & 80.3 & 65.4 \\
API Attack   & \cII 16.1 & \cI 8.2  & \cI 11.6 & \cI 7.5  & \cII 18.2 & \cI 8.5  & \cI 16.4 & \cI 7.0  & \cI 16.1 & \cI 10.2 & \cI 16.8 & \cI 10.4 \\
\midrule
SGLATrack~\cite{sglatrack}    & 84.4 & 65.1 & 84.9 & 66.9 & 81.9 & 59.9 & 81.2 & 62.1 & 82.6 & 65.5 & 82.8 & 67.5 \\
API Attack   & \cI 11.1 & \cII 8.9  & \cII 17.7 & \cII 14.3 & \cI 15.3 & \cII 11.0 & \cII 19.9 & \cII 15.8 & \cII 21.2 & \cII 16.9 & \cII 22.8 & \cII 19.1 \\
\midrule
LGTrack~\cite{LGTrack}      & 84.0 & 65.1 & 83.9 & 65.7 & 80.5 & 58.8 & 87.1 & 65.0 & 82.6 & 64.8 & 80.7 & 65.2 \\
API Attack   & \cIII 47.7 & \cIII 37.6 & \cIII 48.2 & \cIII 38.9 & \cIII 32.5 & \cIII 23.3 & \cIII 44.0 & \cIII 34.5 & \cIII 49.8 & \cIII 39.7 & \cIII 41.9 & \cIII 34.1 \\
\bottomrule
\end{tabular}%
}
\end{table}
\textbf{Benchmarks and Metrics.} To evaluate the API framework, we conduct extensive experiments on six representative UAV tracking benchmarks: DTB70~\cite{dtb70}, UAV123~\cite{uav123} (including the 10fps subset), UAVDT~\cite{uavdt}, VisDrone2018~\cite{2018visdrone}, and UAVtrack112~\cite{uavtrack112}. These datasets collectively encompass a diverse array of aerial challenges, such as intense camera motion, viewpoint variations, and severe occlusions. Performance is quantified using two standard metrics~\cite{OSTrack,fu2022ad}: the Precision Rate (Prec.) and the Success Rate (Succ.). A significant degradation in both indicators relative to the clean baseline signifies a successful adversarial attack.

\subsection{Quantitative Comparison}
To rigorously evaluate the destructive capability of the Adversarial Path-Inversion (API) framework, we conduct a comprehensive comparison against several state-of-the-art adversarial tracking methods. As presented in Table~\ref{tab:attack_performance}, the API framework consistently achieves the most significant degradation in both precision and success rates across all evaluated benchmarks, achieving a remarkable average precision degradation of over 80\% and a success rate drop exceeding 85\%. This uniformity in performance degradation strongly validates the fundamental vulnerability of the dynamic inference topology. For a fair comparison, all baseline attack methods were retrained and evaluated on the AVTrack~\cite{avtrack} model, adhering strictly to the configurations in their respective original settings.

A critical strength of the proposed framework is its ability to balance attack potency with inference efficiency, as detailed in Table~\ref{tab:efficiency_comparison}. Unlike existing adversarial paradigms that typically suffer from an inherent trade-off between performance and speed, the API framework achieves superior results in both dimensions. Specifically, while optimization-based methods such as RTAA~\cite{rtaa} can induce high performance degradation, they are constrained by prohibitive computational overhead, operating at only 4.6 FPS. In contrast, the API framework maintains a high inference speed of 62.5 FPS while delivering an 87.3\% drop in success rate, effectively demonstrating that high-potency attacks do not necessitate a compromise in execution efficiency.

\begin{figure}[!t]
	\centering
	\includegraphics[width=\linewidth]{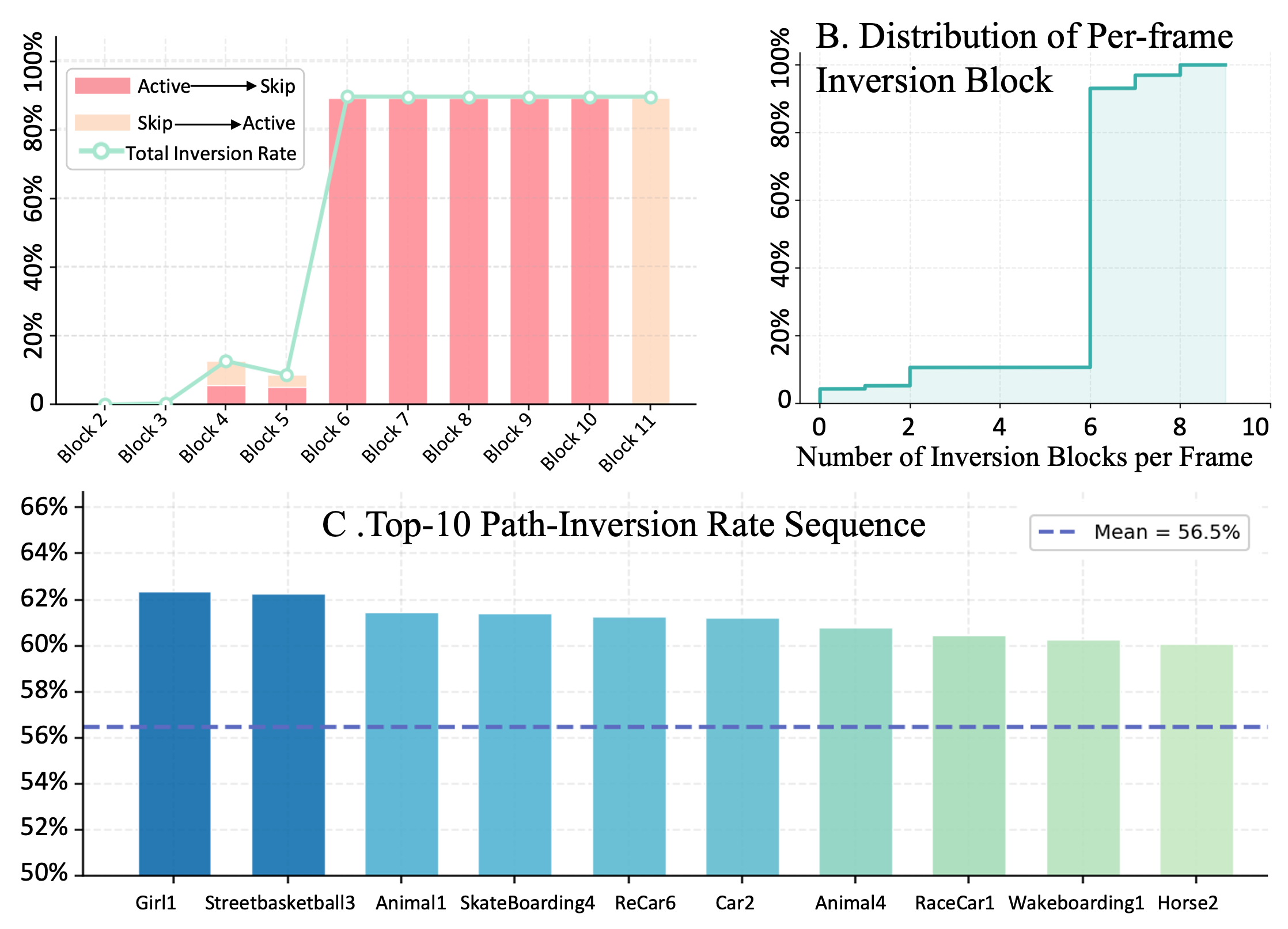}
	\hfill
    \vspace{-2em}
	\caption{Empirical validation of the path-inversion effect across different complementary dimensions.}
	\label{fig:inversion}
    \vspace{-1.5em}
\end{figure}

The generalization capability of the path-based attack is further validated through extensive experiments on three representative adaptive gating dynamic routing trackers: AVTrack~\cite{avtrack}, SGLATrack~\cite{sglatrack}, and LGTrack~\cite{LGTrack}. As shown in Table~\ref{tab:different_model}, the API framework consistently induces substantial performance drops across these diverse architectures, confirming the universal vulnerability of dynamic inference topologies. The weaker degradation on LGTrack is attributable to both a stronger victim baseline and only applies dynamic routing to 3  blocks (30\% of AVTrack's attackable branch space). Consequently, more features are fixed by early blocks, reducing the global impact of route-flip perturbations.

\subsection{Exploration Experiments}
To provide a fine-grained understanding of how the Adversarial Path-Inversion (API) framework operates the dynamic routing topology on AVTrack~\cite{avtrack}, we conduct a systematic exploration of the gate inversion behavior across dimensions, as shown in Fig.~\ref{fig:inversion}.

\textbf{Block-level Analysis.} The observed disparity in path-inversion rates validates the inherent instability of adaptive tracking (see Fig.~\ref{fig:inversion}-A). The sharp phase transition at Block 6 results from the synergy between error propagation and semantic abstraction. In the initial stages (Blocks 2--5), inversion rates remain below 12.7\% as infinitesimal input perturbations have not yet been sufficiently amplified to exceed the decision thresholds of the shallow gating modules. However, through mathematically proven amplification, the cumulative error surpasses the critical threshold of the Lipschitz singularity by the time it reaches the mid-to-deep blocks (Blocks 6--11). This transition is further driven by the high-level semantic representations in deeper layers, which are significantly more susceptible to adversarial feature blurring than the textural features in shallow blocks. Consequently, the inversion rate surges to 89.3\% from Block 6 onward, where original paths are forcibly reversed.

\begin{figure}[!t]
	\centering
	\includegraphics[width=\linewidth]{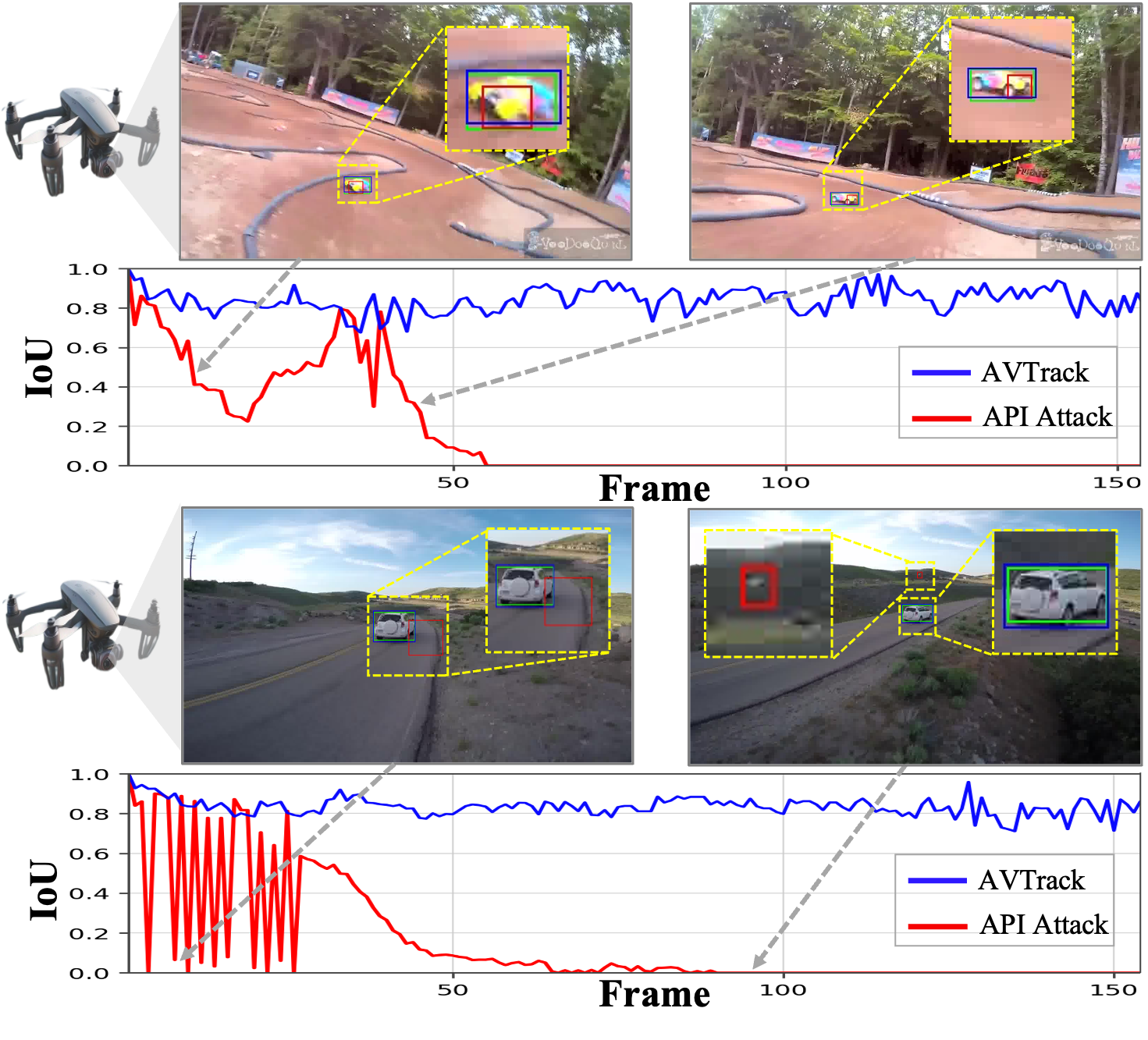}
	\hfill
    \vspace{-3em}
	\caption{Qualitative Visualization of the API Attack on a UAV Tracking Platform.}
    \vspace{-1.2em}
	\label{fig:iou}
\end{figure}
\textbf{Frame-level Effect.} The frame-level distribution in Figure~\ref{fig:inversion}-B  proves this structural interpretation. The distribution exhibits a highly concentrated and near-step transition at 6 inverted blocks, with more than $90\%$ of frames experiencing simultaneous inversion of 6 or more transformer blocks. This near-deterministic behavior confirms that the path-inversion effect is a structurally induced discrete transition: once the perturbation crosses the decision boundary, the routing of multiple deep blocks collapses in unison.

\textbf{Sequence-level Generalizability.} To further assess the stability of the path-inversion effect across diverse aerial tracking scenarios, Figure~\ref{fig:inversion}-C reports the per-sequence gate inversion rate on DTB70, ranked in descending order. The mean inversion rate across the entire benchmark reaches $56.5\%$, with the top-performing sequences consistently exceeding $60\%$.

\subsection{Qualitative Comparison}

As shown in Fig.~\ref{fig:iou}, the corresponding $IoU$ (Intersection over Union) curves below each scenario quantify the tracking precision. Upon the introduction of the adversarial perturbation to the search patch, the adaptive tracker experiences a catastrophic failure, as evidenced by the rapid collapse of the red $IoU$ curve and the resultant irreversible drift of the red bounding box from the actual target.
\begin{figure}[!t]
	\centering
	\includegraphics[width=\linewidth]{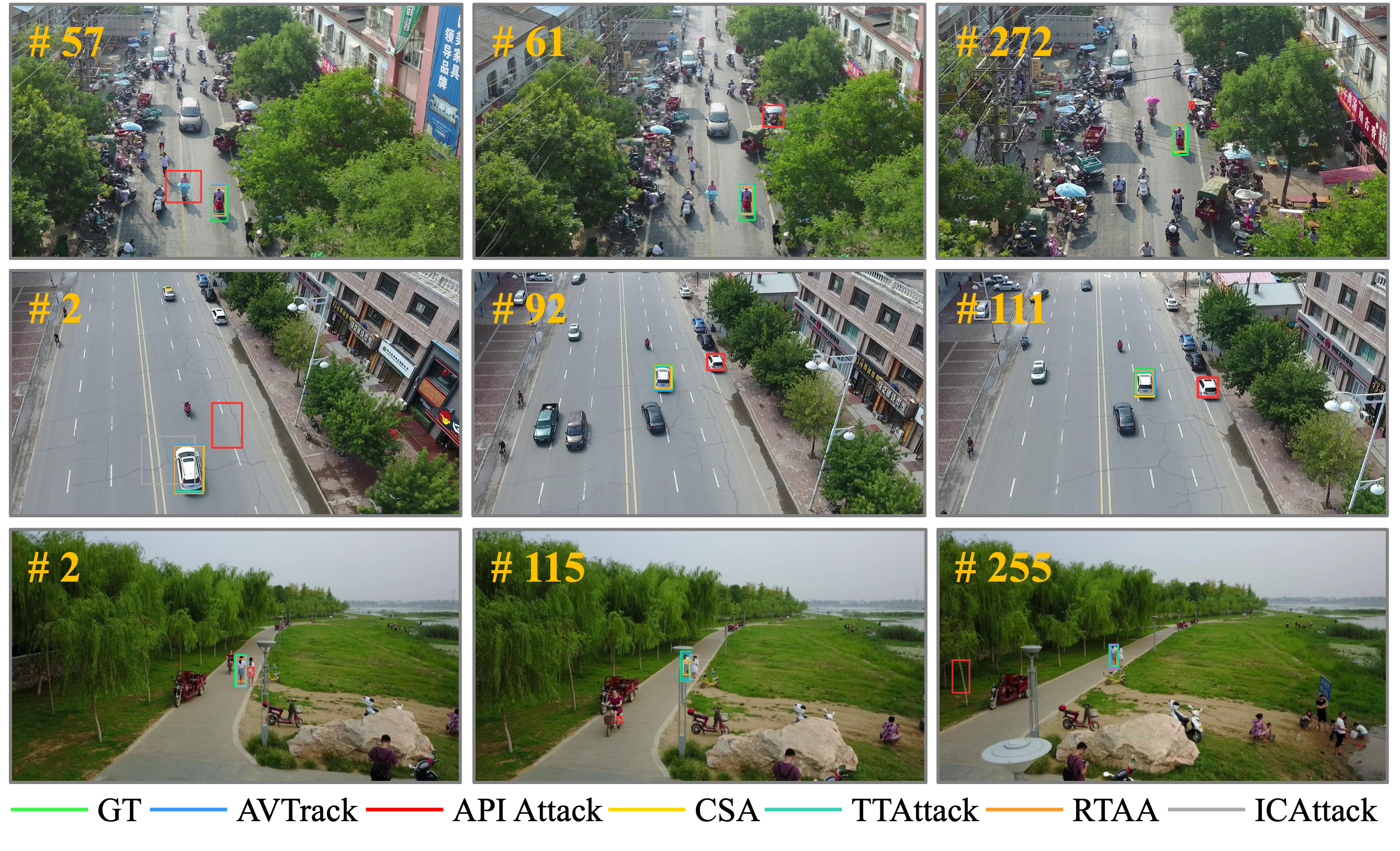}
	\hfill
    \vspace{-2em}
	\caption{Qualitative tracking results of the representative attack methods on the Visdrone2018~\cite{2018visdrone} dataset.}
    \vspace{-1em}
	\label{fig:vis}
\end{figure}

Compared to other adversarial methods, the API framework induces a significantly earlier onset of drift (Fig.~\ref{fig:vis}). For instance, in the middle and bottom rows (at frame \#2), while the baseline adversarial trackers still manage to localize the target, the API attack has already forced the tracker to deviate from the ground truth . Furthermore, the magnitude of the drift generated by the proposed framework is substantially larger than that of the existing methods. As the sequences progress (e.g., frame \#255 and \#272), the $IoU$ ratio of the API-attacked tracker collapses more rapidly and remains consistently lower, eventually approaching zero. This immediate and severe subversion of the tracking logic demonstrates the superior potency of the path-inversion mechanism. 

\begin{figure}[!t]
	\centering
	\includegraphics[width=\linewidth]{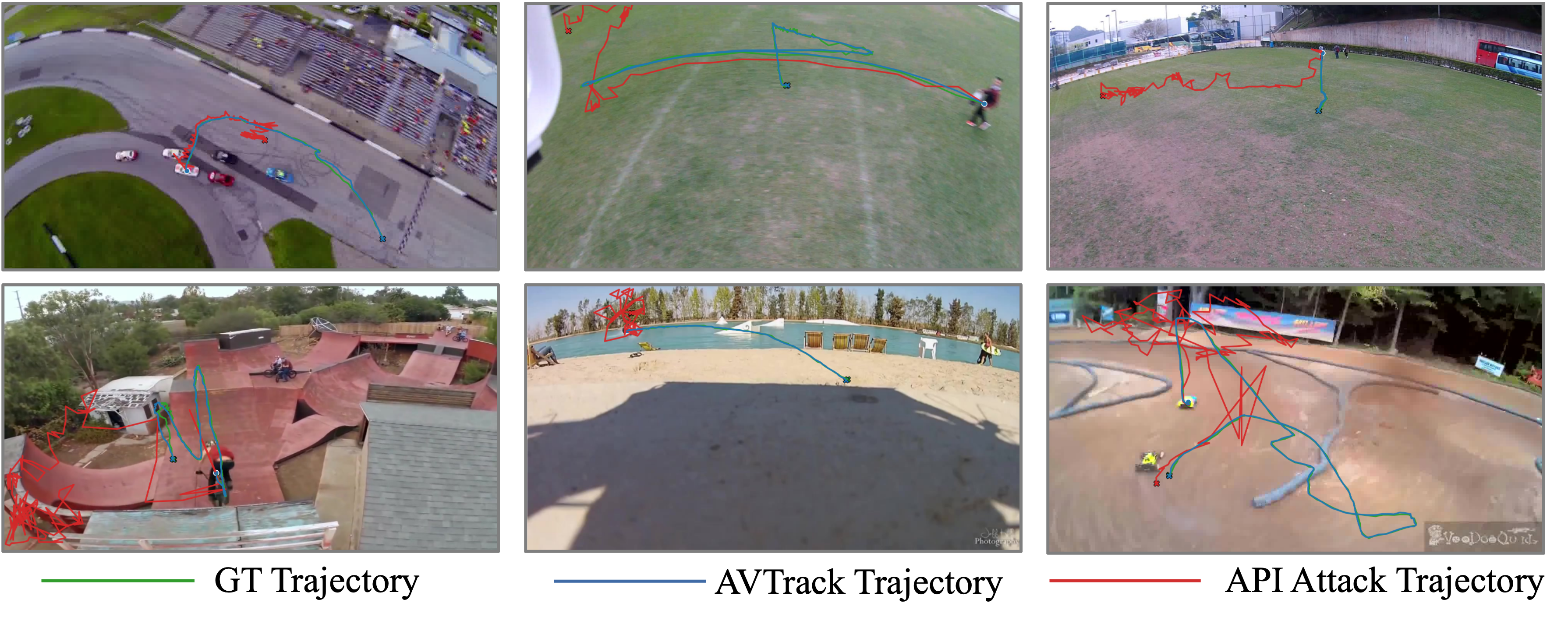}
	\hfill
	\caption{Spatio-temporal trajectory comparison. The attack trajectory deviates early from the baseline path and is repeatedly attracted toward background regions. }
	\label{fig:traj}
\end{figure}

The trajectory visualization jointly explains the attack effectiveness from both temporal and spatial dimensions, as illustrated in Fig.~\ref{fig:traj}. In the temporal dimension, the attack trajectory does not manifest as isolated frame-by-frame noise, but first deviates from the baseline path in the early stage, and then remains continuously unstable in subsequent frames. This pattern indicates the existence of an error accumulation process, where the state update gradually deviates from the true motion manifold and cannot recover. In the spatial dimension, the attack trajectory is repeatedly attracted by background regions and shows frequent local oscillations. Therefore, the proposed attack degrades the instantaneous localization quality and destroys the long-term motion consistency. 

\subsection{Ablation Study}
We conduct comprehensive ablation studies on the DTB70 and UAVDT datasets to dissect the individual contributions of the core components within the Adversarial Path-Inversion framework, as reported in Table~\ref{tab:ablation}. First, we investigate the impact of the path-inversion objective. Removing this component leads to the most significant degradation in attack effectiveness, with the precision score on the DTB70 dataset increasing from 16.1 to 52.7. This severe drop validates the theoretical insight of this paper: without explicitly targeting the Lipschitz singularity to force topological hijacking, relying solely on semantic feature corruption is insufficient to induce catastrophic failure of the tracker.  Second, we analyze the Saliency-Guided Perturbation Focusing module. Disabling this module causes the precision score to recover to 27.9. Without the focusing mechanism, the adversarial energy is uniformly diluted across the background, which reduces the probability of crossing the gating decision boundaries. Furthermore, we evaluate the roles of the learnable noise prior and the noise-to-RGB alignment module. Removing the learnable noise token or the alignment mechanism substantially weakens the attack effectiveness. These results demonstrate that an adaptive noise representation is crucial for bridging the abstract topological manipulation in the latent space to the concrete pixel-level perturbation in the input space. 
The synergy of these components ensures that the structural bias is effectively injected into the search region to dismantle both the localization capability and the motion consistency of the model. More ablation results, additional experimental analysis, and promising defense strategies can be found in the \textbf{supplementary materials}.

\begin{table}[t]
  \centering
  \vspace{-1em}
  \caption{Ablation study of core components on DTB70 and UAVDT datasets.}
  \label{tab:ablation}
  \begin{tabularx}{\linewidth}{@{} l *{4}{>{\centering\arraybackslash}X} @{}}
    \toprule
    \multirow{2}{*}{Variant} & \multicolumn{2}{c}{DTB70} & \multicolumn{2}{c}{UAVDT} \\
    \cmidrule(lr){2-3} \cmidrule(lr){4-5}
    & Succ. & Prec.& Succ. & Prec.\\
    \midrule
    \textbf{Full Model (API)}& \textbf{8.2} & \textbf{16.1} & \textbf{8.5} & \textbf{18.2} \\
    w/o SGPF & 19.8 & 27.9 & 13.5 & 20.1 \\
    w/o Noise2RGB & 21.9 & 30.5 & 17.5 & 26.0 \\
    w/o Path-Inversion & 35.9 & 52.7 & 31.5 & 46.9 \\
    w/o Learn Noise  & 27.9 & 43.2 & 24.5 & 37.7 \\
    \bottomrule
  \end{tabularx}
  \vspace{-1em}
\end{table}

\section{Conclusion}
In this work, we first identify a structural vulnerability in the computation-on-demand mechanism of adaptive tracking architectures. We demonstrate that the discrete gating boundaries of dynamic routing decisions exhibit a Lipschitz singularity. This inherent discontinuity ensures that even infinitesimal input perturbations near the decision boundary can trigger abrupt transitions in the inference topology. To systematically exploit this property, we propose the Adversarial Path-Inversion (API) framework. Different from traditional attack paradigms that target semantic representations within a fixed computational path, our framework operates on a topology-path-based attack surface. By precisely manipulating the gating modules to invert the computational paths, the framework completely dismantles the localization capability and the temporal consistency of the tracker. Extensive evaluations across multiple benchmarks confirm that the proposed method achieves superior attack efficacy and high inference efficiency under strict perturbation constraints. Beyond the attack perspective, this work also motivates the investigation of potential defense strategies, such as randomization of gating decisions, soft relaxation of routing boundaries, and adversarial training with topology-aware regularization. We hope this study provides useful insights for both the security analysis and the robust design of dynamic tracking architectures.

\bibliographystyle{ACM-Reference-Format}
\bibliography{sample-base}
\newpage
\appendix
\section{Appendix}
\subsection{Counterfactual Path Experiment:Causal Decomposition of Attack Effectiveness}
A core claim of this work is that the path-inversion mechanism, rather than conventional semantic feature corruption, constitutes the primary causal factor driving the attack effectiveness. To rigorously validate this claim, we design a counterfactual path experiment that disentangles the contributions of adversarial input perturbation and routing topology alteration.

To isolate the causal contributions of topological transitions versus semantic feature distortion, a causal path analysis is performed across six representative benchmarks for unmanned aerial vehicles (UAVs). This counterfactual analysis involves four distinct experimental configurations: (i) Clean Input with Clean Routing (CC), (ii) Adversarial Input with Adversarial Routing (AA), (iii) Adversarial Input with the original Clean Routing restored (AC), and (iv) Clean Input with the Adversarial Routing injected (CA). Two primary metrics are utilized to quantify these effects: the Recovery Ratio, which measures the fraction of performance degradation reversed by restoring the optimal path, and the Injection Ratio, which quantifies the fraction of damage reproduced by topological manipulation alone, formally:
\begin{equation}
\begin{aligned}
\text{Recovery Ratio} = \frac{\text{AC} - \text{AA}}{\text{CC} - \text{AA}}, \quad
\text{Injection Ratio} = \frac{\text{CC} - \text{CA}}{\text{CC} - \text{AA}}    
\end{aligned}
\end{equation}

The quantitative results summarized in Table~\ref{tab:causal_surgery} demonstrate a catastrophic performance collapse from the \textit{Clean+Clean} baseline (mean Success Rate of 64.5\%) to the \textit{Adv+Adv} state (mean Success Rate of 8.6\%). Notably, in the \textit{Adv+Clean} configuration, the restoration of the optimal computational path facilitates a substantial recovery of tracking precision. For instance, on the DTB70~\cite{dtb70}  dataset, the Success Rate rises from 8.2\% to 50.4\% when the original route is restored to the perturbed input. This significant rebound confirms that the tracker maintains considerable robustness to feature-level noise if the integrity of the inference topology is preserved. Conversely, the \textit{Clean+Adv} configuration reveals that simply altering the routing decisions of a clean input is sufficient to degrade the average Success Rate to 41.68\%.

\begin{table}[h]
\centering
\small 
\caption{Causal path Success Rate results across six UAV benchmarks.}
\vspace{-1em}
\label{tab:causal_surgery}
\renewcommand{\arraystretch}{1.1} 
\begin{tabularx}{\linewidth}{ l *{4}{>{\centering\arraybackslash}X} }
\toprule
\textbf{Dataset} &  \textbf{CC} & \textbf{AA} & \textbf{AC} & \textbf{CA} \\
\midrule
DTB70~\cite{dtb70}  & \cIII 65.0 & \cII 8.2 & \cII 50.4 & \cI \textbf{37.0} \\
UAV123~\cite{uav123} & \cI \textbf{66.8} & 7.5 & 57.8 & 45.4 \\
UAVDT~\cite{uavdt}   & 58.7 & \cI \textbf{8.5} & \cI \textbf{48.0} & \cII 38.5 \\
VisDrone~\cite{2018visdrone}  & 65.0 & \cIII 7.0 & \cIII 55.0 & 43.0 \\
UAV123@10fps~\cite{uav123} & \cII 65.8 & 10.2 & 57.9 & 44.3 \\
UAVTrack112~\cite{uavtrack112}  & 65.4 & 10.4 & 55.9 & \cIII 41.9 \\
\bottomrule
\end{tabularx}
\end{table}
As illustrated in Figure~\ref{fig:causal_path}, the statistical distribution of the ratios across all evaluated benchmarks reinforces the primary role of path inversion in the effectiveness of the attack. The Recovery Ratio remains consistently high, averaging 81.5\%, which indicates that the overwhelming majority of the attack-induced damage originates from the forced transition of the inference topology rather than from semantic distortion. Furthermore, the Injection Ratio averages 40.8\%, proving that route manipulation acts as an independent and potent attack surface. These results establish a strong and consistent causal link between path inversion and attack effectiveness.

\subsection{Depth Amplification Analysis}
The main paper observes a sharp phase transition in path-inversion rates around Block 6 (see Figure 3-A in the main paper). A natural question is how this local effect compounds across the full depth of the network. To systematically characterize this phenomenon, we measure the layer-wise flip rates partitioned into three depth groups in AVTrack~\cite{avtrack}: shallow (Blocks 2--4), middle (Blocks 5--7), and deep (Blocks 8--11).

\begin{table}[h]
\centering
\small 
\caption{Depth-stratified path-inversion rates across four datasets. Deep/Shallow denotes the amplification ratio of the deep-block flip rate relative to the shallow-block rate.}
\vspace{-1em}
\label{tab:depth_stratified}
\renewcommand{\arraystretch}{1.1} 
\begin{tabularx}{\linewidth}{ l *{3}{>{\hsize=0.85\hsize\centering\arraybackslash}X} >{\hsize=1.45\hsize\centering\arraybackslash}X }
\toprule
\textbf{Dataset} & \textbf{Shallow} & \textbf{Middle} & \textbf{Deep} & \textbf{Deep/Shallow} \\
\midrule
DTB70~\cite{dtb70}  & 4.34\% & \cIII 62.42\% & \cIII 89.28\% & \cI \textbf{20.6x} \\
UAV123~\cite{uav123} & \cII 4.68\% & \cI \textbf{64.95\%} & \cI \textbf{92.71\%} & \cIII 19.8x \\
UAVDT~\cite{uavdt}  & \cIII 4.54\% & \cII 64.21\% & \cII 91.48\% & \cII 20.2x \\
VisDrone~\cite{2018visdrone} & \cI \textbf{5.70\%} & 61.37\% & 85.95\% & 15.1x \\
\midrule
\rowcolor{gray!15} 
\textbf{Mean} & \textbf{4.82\%} & \textbf{63.24\%} & \textbf{89.85\%} & \textbf{18.9x} \\
\bottomrule
\end{tabularx}
\end{table}

The mean deep-block flip rate reaches 89.85\%, nearly 19 times higher than the shallow-block rate of 4.82\%. This extreme amplification ratio confirms that the path-inversion effect exhibits a strongly nonlinear depth dependency. The underlying mechanism can be understood as a cascade: infinitesimal input perturbations are insufficient to cross the gating thresholds of early blocks (shallow flip rate below 5\%), but each flipped gate in the intermediate layers alters the feature representation passed to subsequent layers, progressively amplifying the effective perturbation magnitude. By the time the signal reaches the deep blocks, the cumulative feature distortion has grown large enough to overwhelm the gating thresholds, resulting in near-total inversion (89.85\%).

\section{Temporal and Distributional Analysis of Attack Effectiveness}
\label{sec:supp_temporal}
\subsection{Temporal Damage Persistence Analysis}
To quantify the temporal dynamics of the attack, we measure the Time-to-Failure (TTF), defined as the first frame at which the IoU between the predicted and ground-truth bounding boxes falls below 0.5, as shown in Table~\ref{tab:ttf_analysis}. A lower TTF indicates faster failure onset.
\begin{table}[t]
\centering
\small 
\caption{Time-to-Failure analysis across six benchmarks. TTF Ratio denotes the ratio of adversarial TTF to clean TTF; smaller values indicate faster attack-induced failure.}
\vspace{-1em}
\label{tab:ttf_analysis}
\renewcommand{\arraystretch}{1.1} 
\begin{tabularx}{\linewidth}{ l *{4}{>{\centering\arraybackslash}X} }
\toprule
\textbf{Dataset} & \textbf{$N$} & \textbf{TTF (Clean)} & \textbf{TTF (Adv+Adv)} & \textbf{TTF Ratio} \\
\midrule
DTB70~\cite{dtb70}  & 70 & 175.63 & \cI \textbf{24.84} & 0.141 \\
UAV123~\cite{uav123} & 123 & \cI \textbf{765.85} & 95.07 & \cIII 0.124 \\
UAVDT~\cite{uavdt}  & 50 & 598.06 & 74.74 & 0.125 \\
VisDrone~\cite{2018visdrone} & 35 & \cII 743.77 & 71.20 & \cI \textbf{0.096} \\
UAV123@10fps~\cite{uav123} & 123 & 259.43 & \cII 47.55 & 0.183 \\
UAVTrack112~\cite{uavtrack112} & 112 & \cIII 641.54 & \cIII 68.57 & \cII 0.107 \\
\midrule
\rowcolor{gray!15} 
\textbf{Mean} & & & & \textbf{0.129} \\
\bottomrule
\end{tabularx}
\end{table}

The mean TTF ratio of 0.129 indicates that the API attack causes tracking failure approximately 7.8 times earlier than the natural failure point of the clean tracker. On VisDrone, the ratio drops to 0.096, meaning the attack induces failure more than 10 times faster. This rapid onset is consistent with the cascade amplification mechanism described in Section F: once the adversarial perturbation triggers route inversion in the early-to-middle blocks, the resulting feature inconsistency propagates and accumulates through subsequent layers, leading to rapid and irreversible tracking drift.

\begin{table}[t]
\centering
\small 
\caption{Recovery analysis after route restoration.}
\vspace{-1em}
\label{tab:recovery_analysis}
\renewcommand{\arraystretch}{1.1} 
\begin{tabularx}{\linewidth}{ l *{3}{>{\centering\arraybackslash}X} }
\toprule
\textbf{Dataset} & \textbf{TTF (Adv+Adv)} & \textbf{TTF (Adv+Clean)} & \textbf{Recovery Gain (IoU)} \\
\midrule
DTB70~\cite{dtb70}  & \cI \textbf{24.84} & 144.90 & \cIII 0.432 \\
UAVDT~\cite{uavdt}  & 74.74 & \cIII 472.10 & 0.408 \\
VisDrone~\cite{2018visdrone} & \cIII 71.20 & \cI \textbf{564.14} & \cI \textbf{0.479} \\
UAVTrack112~\cite{uavtrack112} & \cII 68.57 & \cII 534.73 & \cII 0.473 \\
\bottomrule
\end{tabularx}
\end{table}

To assess whether the temporal damage is persistent or reversible, we measure the TTF when the adversarial routing decisions are replaced with clean routing decisions while retaining the adversarial input (the Adv+Clean condition from the path experiment).

\begin{figure}[t]
	\centering
	\includegraphics[width=\linewidth]{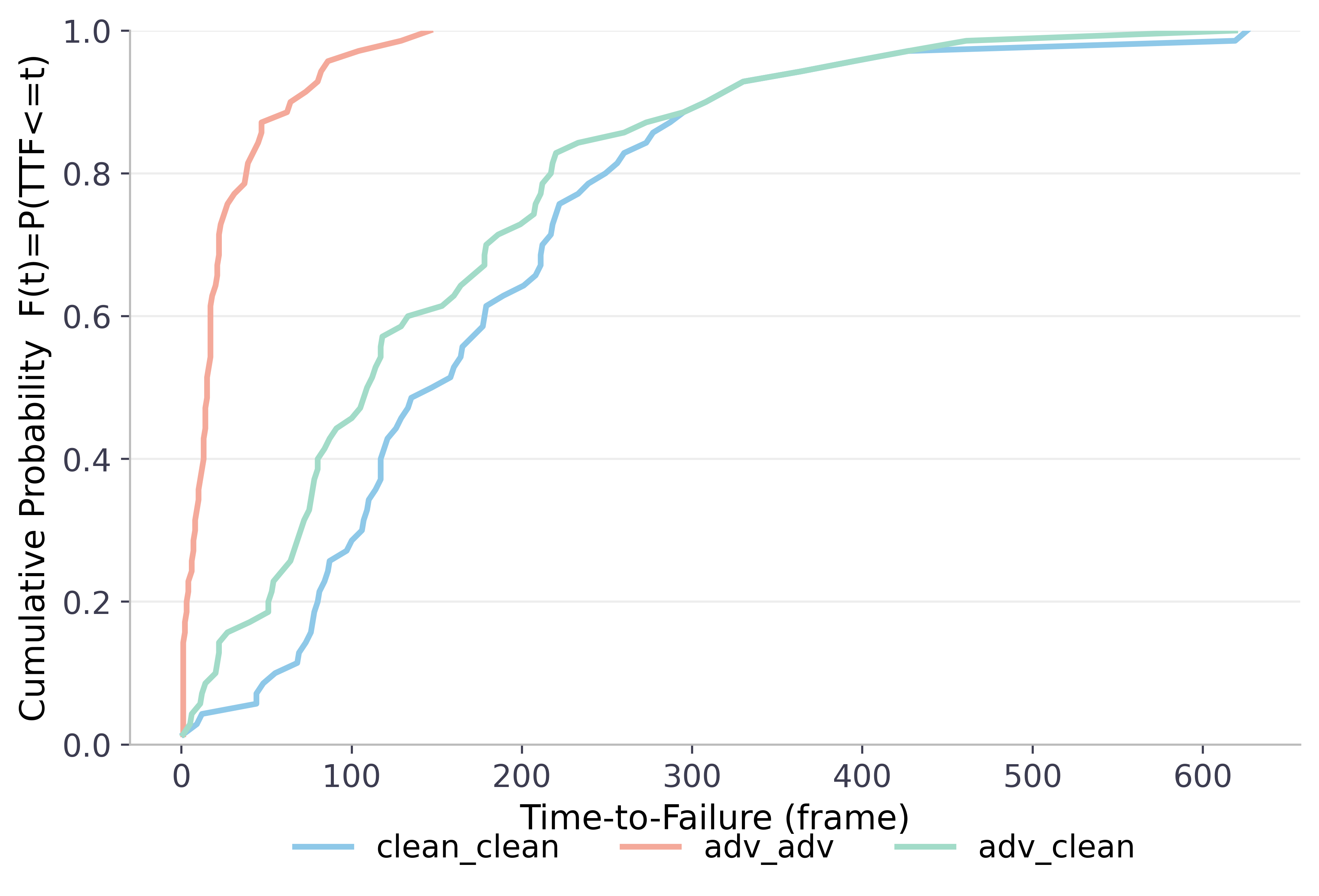}
	\hfill
    \vspace{-2em}
	\caption{Time-to-Failure cumulative distribution on DTB70~\cite{dtb70} .}
	\label{fig:ttf}
    \vspace{-1em}
\end{figure}
\begin{figure}[t]
	\centering
	\includegraphics[width=\linewidth]{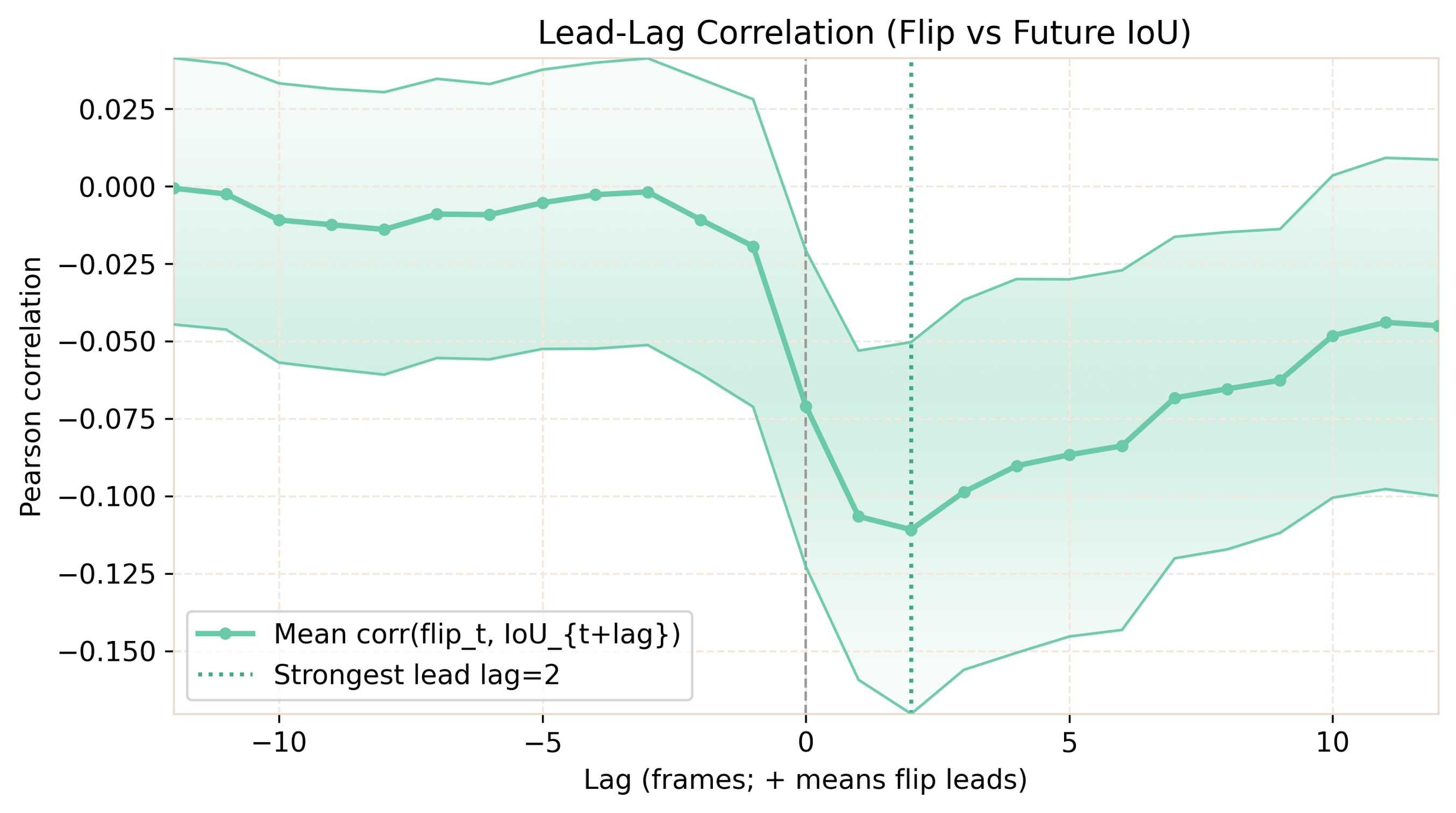}
	\hfill
    \vspace{-2em}
	\caption{The path flip and frame IoU temporal causality analysis.}
	\label{fig:lead_frame}
    \vspace{-1em}
\end{figure}
Restoring the clean routing decisions substantially extends TTF by a factor of 5--8$\times$ (Table~\ref{tab:recovery_analysis}), confirming that temporal collapse is dominated by topological path alteration rather than input-level semantic corruption. This trend is also consistent with the rightward shift of the Adv+Clean curve relative to Adv+Adv in Figure~\ref{fig:ttf}. However, the Adv+Clean TTF remains below the clean baseline TTF (e.g., 144.90 vs. 175.63 on DTB70~\cite{dtb70} ), indicating that a residual component of the temporal damage originates from the adversarial input perturbation itself. 

The sharp leftward shift of the adv\_adv curve (pink) relative to the clean\_clean curve (blue) demonstrates rapid failure onset. The adv\_clean curve (green) lies between the two, confirming that route restoration substantially delays failure while the residual adversarial input effect prevents full recovery.

Temporal analysis in Figure~\ref{fig:lead_frame} shows that routing instability tends to appear before tracking collapse. 
The strongest negative lead-lag correlation is observed at a positive lag of two frames, 
$ corr(\mathrm{flip}_t,\mathrm{IoU}_{t+\mathrm{lag}})=-0.1109 $. 
This means that an increase of route-flip rate at frame $t$ is typically followed by an IoU drop around frame $t+2$. 
Compared with lag zero, the negative-correlation gain is 0.0368, and the 95\% confidence interval is strictly positive, 
which indicates that this two-frame lead effect is stable rather than random noise.

\section{Extended Qualitative Comparison}
\label{Qualitative}
\begin{figure}[t]
	\centering
	\includegraphics[width=\linewidth]{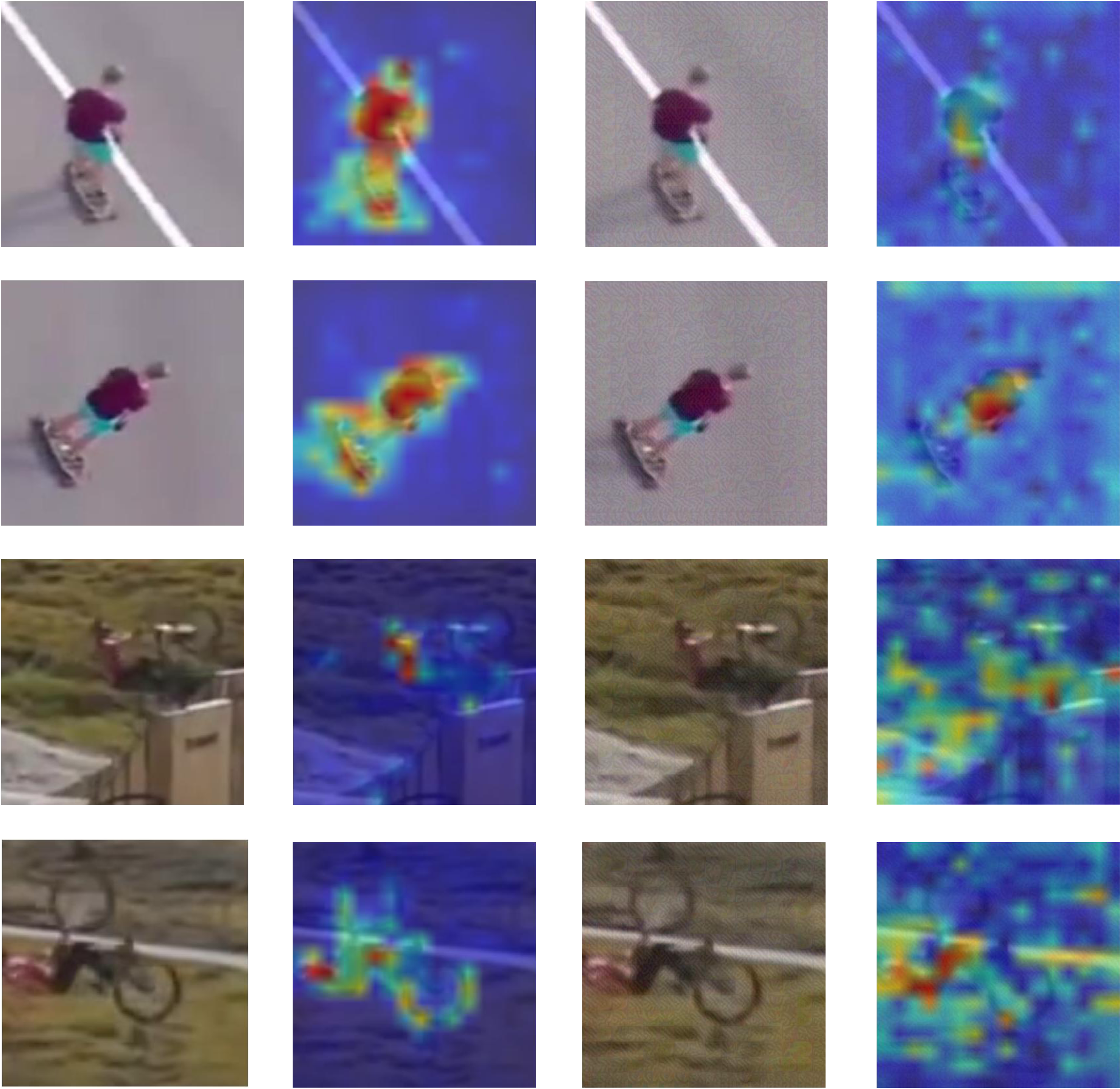}
	\hfill
    \vspace{-2em}
	\caption{Comparison of feature activation heatmaps for clean and perturbed inputs.}
	\label{fig:heatmap}
    \vspace{-1em}
\end{figure}

Figure~\ref{fig:heatmap} compares activation heatmaps before and after API intervention, where the left panel in each pair corresponds to the clean input and its activation heatmap, and the right panel corresponds to the perturbed input and its heatmap under API attack. For clean inputs, high-response regions are compact and target-centered, indicating well-localized and discriminative attention. After perturbation, activations become spatially diffuse, with increased background responses and reduced target–background contrast. This attention dispersion suggests that API primarily disrupts the model’s spatial selectivity, causing the tracker to rely on unstable or spurious cues and thereby increasing drift and failure risk.

\begin{figure*}[t]
	\centering
	\includegraphics[width=\linewidth]{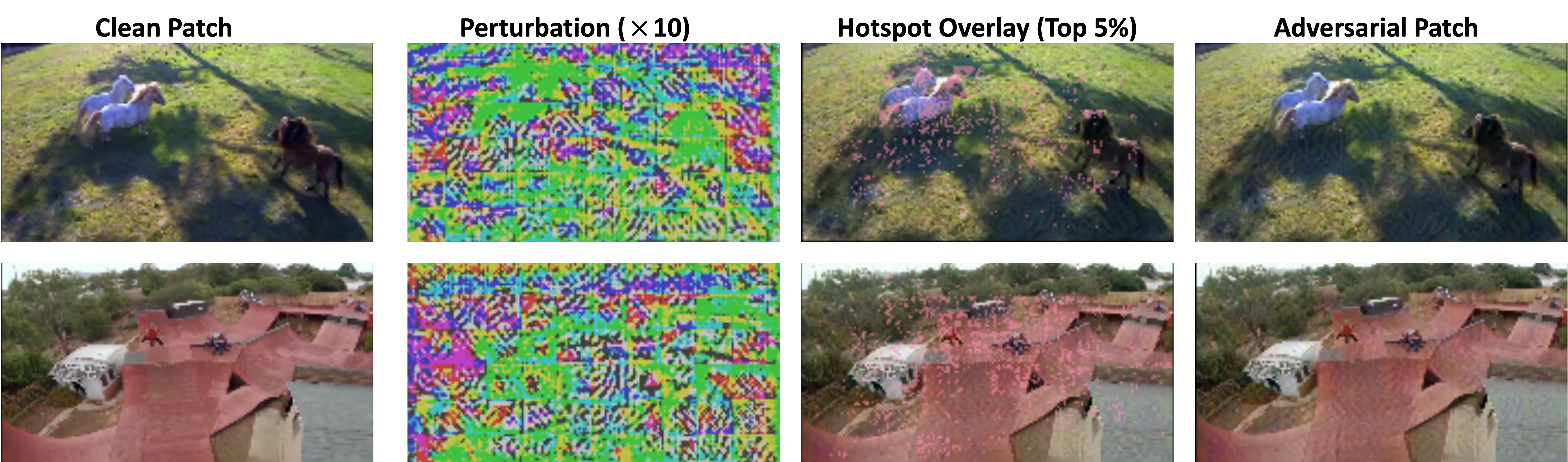}
	\hfill
    \vspace{-2em}
	\caption{Spatial distribution of API perturbations on representative search
patches. From left to right, the columns show the clean patch, the
perturbation amplified by $10\times$ for visualization, the overlay of
the top 5\% perturbation hotspots, and the resulting adversarial patch.}
	\label{fig:vis}
    \vspace{-1em}
\end{figure*}
Figure~\ref{fig:vis} presents a four-column visualization pipeline for representative search patches. The first column shows the clean input patch. The second column shows the perturbation amplified by a factor of ten, which is used only for visualization because the original perturbation magnitude is small. The third column shows the Top 5\% hotspot overlay, which denotes the most strongly perturbed spatial locations in each patch. The fourth column shows the final adversarial patch after perturbation injection. For the horse and rooftop examples, the Top 5\% hotspots are mainly concentrated around the target body and structural boundaries, while large background regions remain sparse. This pattern indicates strong spatial selectivity of API perturbations. The perturbation budget is preferentially allocated to regions that are more relevant to tracking decisions, rather than being spread as global destructive noise. 

\section{Extended Ablation Studies}
\label{sec:supp_ablation}
\subsection{Loss-Term Sensitivity Analysis}
To complement Table~4 in the main paper, we provide additional parameter-level ablations on DTB70~\cite{dtb70}  and UAVDT~\cite{uavdt}  under the same perturbation budget ($\ell_\infty \leq 6.5/255$). As in the main paper, lower Success and Precision indicate stronger attack effectiveness.

\begin{table}[t]
\centering
\small
\caption{Loss-term sensitivity and objective-removal ablation under matched-budget evaluation.}
\vspace{-1em}
\label{tab:loss_sensitivity_formatted}
\renewcommand{\arraystretch}{1.1}
\begin{tabularx}{\linewidth}{ l *{4}{>{\centering\arraybackslash}X} }
\toprule
\multirow{2}{*}{\textbf{$(\lambda_{\text{recon}},\lambda_{\text{resp}},\lambda_{\text{path}},\lambda_{\text{feat}})$}} & \multicolumn{2}{c}{\textbf{DTB70~\cite{dtb70} }} & \multicolumn{2}{c}{\textbf{UAVDT~\cite{uavdt} }} \\
\cmidrule(lr){2-3} \cmidrule(lr){4-5}
 & \textbf{Succ.} & \textbf{Prec.} & \textbf{Succ.} & \textbf{Prec.} \\
\midrule
Full API & \cI \textbf{10.11} & \cI \textbf{19.59} & \cI \textbf{9.20} & \cI \textbf{20.57} \\
$(1,10^3,1,1)$ & \cIII 14.60 & \cIII 23.62 & \cIII 14.79 & \cIII 29.79 \\
$(1,10^4,0.1,1)$ & 65.61 & 84.82 & 55.19 & 74.19 \\
$(1,10^4,5,1)$ & \cII 11.65 & \cII 22.66 & \cII 12.04 & \cII 27.00 \\
$(1,10^4,1,0.1)$ & 26.11 & 41.40 & 23.05 & 36.42 \\
$(0.1,10^4,1,1)$ & 22.68 & 34.48 & 20.01 & 31.88 \\
\bottomrule
\end{tabularx}
\end{table}
Table~\ref{tab:loss_sensitivity_formatted} shows a clear hierarchy of objective importance. Comparing Row 2 (Full API) with Row 4 ($\lambda_{\text{path}}=0.1$), the attack degrades most severely, indicating that the path-inversion term is the primary driver of effectiveness. Comparing Row 2 with Row 5 ($\lambda_{\text{path}}=5$), the change is much smaller, suggesting that the baseline gate weight is already in a near-effective operating region rather than under-regularized.
Rows 3, 6, and 7 further indicate that response suppression, feature-space disruption, and reconstruction-constrained generation are all necessary to stabilize the attack pipeline. In particular, weakening $\lambda_{\text{resp}}$ (Row 3) consistently reduces attack strength, while weakening $\lambda_{\text{feat}}$ (Row 6) or $\lambda_{\text{recon}}$ (Row 7) also causes clear performance recovery of the victim tracker. Taken together, the table supports a mechanism-level conclusion: topology inversion is the dominant component, and the other terms mainly act as complementary constraints that preserve transferability and optimization stability.

\begin{table}[t]
\centering
\small
\caption{Objective-term ablation for $L_{\text{resp}}$ and $L_{\text{feat}}$. Lower is better for attack effectiveness.}
\label{tab:loss_term_ablation}
\renewcommand{\arraystretch}{1.1}
\begin{tabularx}{\linewidth}{ l *{4}{>{\centering\arraybackslash}X} }
\toprule
\multirow{2}{*}{\textbf{Setting}} & \multicolumn{2}{c}{\textbf{DTB70~\cite{dtb70} }} & \multicolumn{2}{c}{\textbf{UAVDT~\cite{uavdt} }} \\
\cmidrule(lr){2-3} \cmidrule(lr){4-5}
 & \textbf{Succ.} & \textbf{Prec.} & \textbf{Succ.} & \textbf{Prec.} \\
\midrule
Full API & \cI \textbf{8.20} & \cI \textbf{16.10} & \cI \textbf{8.50} & \cI \textbf{18.20} \\
w/o $L_{\text{feat}}$ & \cII 9.78 & \cII 17.50 & \cII 9.30 & \cII 20.23 \\
w/o $L_{\text{recon}}$ & \cIII 13.82 & \cIII 29.81 & 20.47 & 43.37 \\
w/o $L_{\text{resp}}$  & 16.64 & 34.03 & \cIII 19.28 & \cIII 40.53 \\
\bottomrule
\end{tabularx}
\end{table}

Table~\ref{tab:loss_term_ablation} indicates that removing $L_{\text{resp}}$ causes a large recovery of victim performance on both datasets, while removing $L_{\text{feat}}$ leads to a much smaller change. Removing $L_{\text{resp}}$ causes substantial recovery of victim performance on both datasets (DTB70~\cite{dtb70} : 16.64/34.03; UAVDT~\cite{uavdt} : 19.28/40.53), whereas removing $L_{\text{feat}}$ leads to marginal change under the current configuration. This pattern suggests that $L_{\text{resp}}$ is indispensable, while the marginal utility of $L_{\text{feat}}$ is weaker.

\subsection{SGPF Internal Parameter Ablation}
In the API generator, Saliency-Guided Perturbation Focusing (SGPF) computes a binary top-$K$ token mask $W$ from attention scores, and then forms token-centric adversarial features as:
\begin{equation}
    T^{tc} = W \odot (T_s + T_n) + \gamma (1-W)\odot T_n.
\end{equation}

Therefore, $\gamma$ controls how much perturbation energy is retained on non-salient tokens: $\gamma=1.0$ is uniform allocation, while $\gamma=0.0$ is hard suppression on non-salient regions.

\begin{table}[t]
\centering
\small
\caption{SGPF internal parameter ablation. Lower is better for attack effectiveness.}
\label{tab:sgpf_gamma_ablation}
\renewcommand{\arraystretch}{1.1}
\begin{tabularx}{\linewidth}{ l *{4}{>{\centering\arraybackslash}X} }
\toprule
\multirow{2}{*}{\textbf{Setting}} & \multicolumn{2}{c}{\textbf{DTB70~\cite{dtb70} }} & \multicolumn{2}{c}{\textbf{UAVDT~\cite{uavdt} }} \\
\cmidrule(lr){2-3} \cmidrule(lr){4-5}
 & \textbf{Succ.} & \textbf{Prec.} & \textbf{Succ.} & \textbf{Prec.} \\
\midrule
SGPF with $\gamma=0.5$ (API) & \cI \textbf{8.20} & \cI \textbf{16.10} & \cI \textbf{8.50} & \cI \textbf{18.20} \\
SGPF with $\gamma=1.0$  & 24.49 & 36.01 & 24.38 & 39.76 \\
SGPF with $\gamma=0.0$ & 25.26 & \cIII 34.04 & \cIII 20.49 & \cIII 34.16 \\
w/o SGPF & \cII 19.80 & \cII 27.90 & \cII 13.50 & \cII 20.10 \\
\bottomrule
\end{tabularx}
\end{table}

Table~\ref{tab:sgpf_gamma_ablation} shows that SGPF is not only necessary, but also sensitive to internal energy-allocation balance. The API default ($\gamma=0.5$) consistently gives the strongest attack, while both boundary settings ($\gamma=1.0$ and $\gamma=0.0$) cause clear victim-performance recovery. This indicates that neither uniform perturbation spreading nor overly hard background suppression is optimal for path-inversion transfer.
The comparison with \textit{w/o SGPF} further suggests that SGPF contributes beyond a simple masking operation: its value lies in calibrated redistribution of perturbation energy between salient and non-salient tokens. In other words, $\gamma=0.5$ acts as a structural compromise that preserves target-directed disruption while maintaining sufficient global coupling for stable attack propagation.

\begin{figure}[t]
	\centering
	\includegraphics[width=\linewidth]{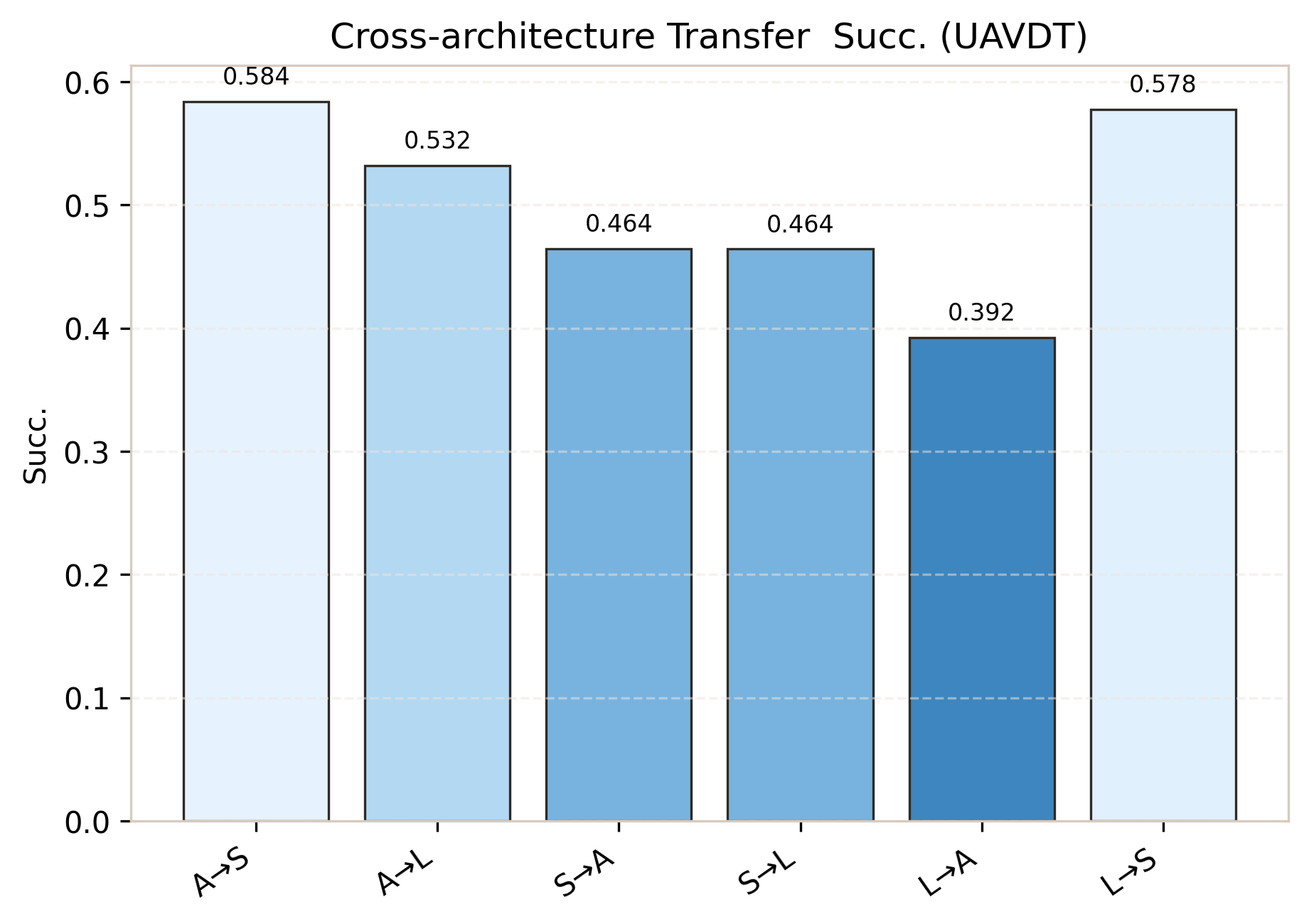}
	\hfill
    \vspace{-2em}
	\caption{Cross-architecture transfer performance on UAVDT. $A$, $S$, and $L$ denote AVTrack, SGLATrack, and LGTrack, respectively.}
	\label{fig:cross_attack}
    \vspace{-1em}
\end{figure}
\subsection{Cross-Architecture Transfer Analysis}
The results in Figure~\ref{fig:cross_attack} show a clear asymmetry between matched and mismatched source--target settings. Under matched settings, the attack remains strong, for example, the attacked scores on UAVDT~\cite{uavdt}  are reduced to $18.2\%/8.5\%$ for AVTrack, $15.3\%/11.0\%$ for SGLATrack, and $32.5\%/23.3\%$ for LGTrack in Prec./Succ., respectively. Under cross-architecture transfer, the degradation is weaker and heterogeneous. This behavior indicates that transfer is present but limited, rather than uniformly strong across all architecture pairs.
This pattern is consistent with the proposed route-singularity mechanism. The optimization of perturbation is guided by local gate decision geometry, while the geometry of routing boundaries depends on tracker architecture, including gate parameterization, branch semantics, token pipeline, and normalization. As a consequence, a perturbation optimized for one tracker does not align perfectly with the routing boundaries of another tracker. The observed gap is therefore a mechanism-level property, not an implementation inconsistency.

\subsection{Migration of Similar Dynamic Routing Attack Methods}
\begin{table}[t]
\centering
\small
\caption{Comparison on UAV123~\cite{uav123} and UAVDT~\cite{uavdt}  between clean tracking, API attack, and SlowFormer transfer attack. }
\label{tab:slowformer_vs_api_uav}
\renewcommand{\arraystretch}{1.1}
\begin{tabularx}{\linewidth}{l *{4}{>{\centering\arraybackslash}X}}
\toprule
\multirow{2}{*}{\textbf{Method}} & \multicolumn{2}{c}{\textbf{UAV123~\cite{uav123}}} & \multicolumn{2}{c}{\textbf{UAVDT~\cite{uavdt} }} \\
\cmidrule(lr){2-3} \cmidrule(lr){4-5}
 & \textbf{Prec.} & \textbf{Succ.} & \textbf{Prec.} & \textbf{Succ.} \\
\midrule
AVTrack (Clean) & 84.8 & 66.8 & 82.1 & 58.7 \\
SlowFormer (Transfer) & 82.7& 65.3 & 78.0 & 55.9 \\
API Attack & \textbf{11.6} & \textbf{7.5} & \textbf{18.2} & \textbf{8.5} \\
\bottomrule
\end{tabularx}
\end{table}

To provide a clearer method-level comparison with existing dynamic-routing attacks, we attempted to transfer two representative approaches from Section 2.2 (SlowFormer~\cite{slowformer} and DeepSloth~\cite{Slowdown}) into our tracking framework. The transfer results show that the SlowFormer-style attack is implementable but yields only limited degradation in tracking quality, with performance remaining close to clean AVTrack and substantially weaker than API Attack (Table~\ref{tab:slowformer_vs_api_uav}). 
DeepSloth, in contrast, cannot be migrated as is in a methodologically equivalent way. Its objective is structurally defined on multi-exit SDN classifiers and requires intermediate-exit logits and explicit early-exit decision variables. AVTrack is a single-head localization model (score/size/offset outputs) and does not expose homologous intermediate classification exits. Therefore, the original DeepSloth objective is not identifiable on AVTrack’s computation graph; the issue is objective mismatch rather than hyperparameter tuning.

More importantly, our distinction from~\cite{slowformer,Slowdown} is mechanistic, not merely task-level. Prior methods primarily optimize efficiency-related targets (e.g., delayed exits or increased active computation), whereas API explicitly optimizes \emph{path inversion} in binary routing and couples it with localization-response collapse. API is formulated from the Lipschitz-singularity perspective of discrete routing boundaries and supported by causal path-surgery evidence, which explains why it produces much stronger attack effectiveness in adaptive tracking.
\end{document}